\documentclass{article}
\PassOptionsToPackage{table}{xcolor}
\usepackage[final]{colm2026_conference}
\usepackage{microtype}
\usepackage{graphicx}
\usepackage{subcaption}
\usepackage{booktabs}
\usepackage{hyperref}
\usepackage{url}
\usepackage{multirow}

\usepackage[T1]{fontenc}
\usepackage[utf8]{inputenc}

\usepackage[most]{tcolorbox}
\usepackage{listings}
\usepackage{upquote}
\usepackage{soul}
\usepackage{xcolor}

\usepackage{lineno}
\usepackage{wrapfig}

\definecolor{darkblue}{rgb}{0, 0, 0.5}
\hypersetup{colorlinks=true, citecolor=darkblue, linkcolor=darkblue, urlcolor=darkblue}

\lstdefinestyle{promptstyle}{
	basicstyle=\ttfamily\footnotesize,
	breaklines=true,
	columns=fullflexible,
	keepspaces=true,
	showstringspaces=false,
	upquote=true,
	literate=
	{≤}{{\ensuremath{\le}}}1
	{≥}{{\ensuremath{\ge}}}1
	{-}{{---}}1
	{-}{{--}}1
	{...}{{\ldots}}1
	{"}{{``}}1
	{"}{{''}}1
	{'}{{'}}1
	{•}{{$\bullet$ }}1
}

\lstdefinestyle{pseudostyle}{
	basicstyle=\ttfamily\scriptsize,
	breaklines=true,
	columns=fullflexible,
	keepspaces=true,
	showstringspaces=false,
	upquote=true,
	morecomment=[l]{\#},
	commentstyle=\color{black!55}\itshape
}

\usepackage{amsmath}
\usepackage{amssymb}
\usepackage{mathtools}
\usepackage{amsthm}

\usepackage[capitalize,noabbrev]{cleveref}

\theoremstyle{plain}

\theoremstyle{definition}

\theoremstyle{remark}

\title{DeepThinkVLA: Enhancing Reasoning Capability of Vision-Language-Action Models}

\author{Cheng Yin$^{1,5,6}$, Yankai Lin$^{2,3}$\thanks{Corresponding authors}, Wang Xu$^{4}$, Sikyuen Tam$^{4}$, \textbf{Xiangrui Zeng}$^{1,5*}$, \\\textbf{Zhiyuan Liu}$^{4}$, \textbf{Zhouping Yin}$^{1,5}$ \\
$^1$Research Center for Advanced Electronics Manufacturing, School of Mechanical \\Science and Engineering, Huazhong University of Science and Technology\\
$^2$Gaoling School of Artificial Intelligence, Renmin University of China\\
$^3$Beijing Academy of Artificial Intelligence\\
$^4$Department of Computer Science and Technology, Tsinghua University\\
$^5$State Key Laboratory of Intelligent Manufacturing Equipment and Technology, \\Huazhong University of Science and Technology\\
$^6$Beijing Zhongguancun Academy\\
\texttt{yinchenghust@hust.edu.cn}, \texttt{yankailin@ruc.edu.cn}, \texttt{zeng@hust.edu.cn}
}

\begin{document}

\ifcolmsubmission
\linenumbers
\fi

\maketitle

\begin{abstract}
  Does Chain-of-Thought (CoT) reasoning genuinely improve Vision--Language--Action (VLA) models, or does it merely add overhead? Existing CoT-VLA systems report limited and inconsistent gains, yet no prior work has rigorously diagnosed when and why CoT helps robots act. Through systematic experiments, we identify two necessary conditions that must be jointly satisfied for CoT to be effective in VLA: (1)~\textbf{Decoding Alignment}---CoT and actions must be generated with modality-appropriate mechanisms; forcing both through a single autoregressive decoder is not merely suboptimal but actively harmful, degrading performance by 4.2 percentage points; (2)~\textbf{Causal Alignment}---CoT must be causally linked to task success via outcome-based optimization; without it, supervised CoT is indistinguishable from no reasoning at all under action-execution-sensitive dynamics shift, exhibiting a 32.0\,pp performance drop nearly identical to the 31.6\,pp drop of a reasoning-free baseline. Guided by these findings, we build DeepThinkVLA: a hybrid-attention decoder satisfies Condition~1 by pairing causal attention for language with bidirectional attention for parallel action decoding, while a two-stage SFT-then-RL pipeline satisfies Condition~2 by aligning the full reasoning--action chain with sparse task-success rewards. DeepThinkVLA achieves 97.0\% success on LIBERO, 79.0\% robustness on LIBERO-Plus (vs.\ 61.6\% for $\pi_0$\texttt{-FAST}), and 59.3\% success on RoboTwin~2.0, exceeding the strongest baseline by 21.7 points. Furthermore, real-robot experiments provide preliminary evidence for the physical applicability of our CoT data construction and hybrid architecture. Our codes are available at \url{https://github.com/OpenBMB/DeepThinkVLA}.
\end{abstract}

\section{Introduction}

    Vision--Language--Action (VLA) models have driven notable progress in robotic manipulation, enabling tasks such as stacking blocks, opening drawers, and arranging household objects~\citep{huang2022inner, zitkovich2023rt, yang2024robot, cadene2024lerobot}.
    The dominant paradigm learns a reactive, end-to-end policy that directly maps high-level goals and sensory inputs to low-level motor commands~\citep{chi2023diffusion, kim2024openvla, bjorck2025gr00t}.
    While effective in standard settings, these System-1-style reactive policies~\citep{cui2025openhelix} tend to overfit to training dynamics and struggle to generalize under environmental perturbations or out-of-distribution (OOD) conditions~\citep{ma2024survey, liu2025aligning}.

    A natural remedy is to endow VLAs with the capacity to ``think before acting'' via Chain-of-Thought (CoT) reasoning~\citep{zawalski2024robotic, chen2025training}.
    Recent work takes initial steps in this direction by fine-tuning VLAs on CoT-annotated embodied data~\citep{zawalski2024robotic, lin2025onetwovla, tan2025reason}.
    The premise is appealing: explicit reasoning should decompose the monolithic perception-to-action mapping, thereby bridging the semantic gap and improving generalization. In practice, however, existing CoT-VLA systems report marginal and highly task-dependent gains~\citep{zhao2025cot, liu2025spatialcot}.
    This raises a fundamental, open question: Does CoT reasoning genuinely participate in robotic decision-making, or is it mere decoration---plausible text that adds latency without improving actions?
    
    In this work, we provide a systematic answer. Through controlled experiments, we identify two necessary conditions that must be jointly satisfied for CoT to be effective in VLA models---and show that violating either renders CoT useless or actively harmful. 
    
    \textbf{(1)~Condition~1: Decoding Alignment.} CoT and actions must be generated with modality-appropriate decoding mechanisms. 
    Language reasoning is inherently sequential and suits autoregressive generation~\citep{NonAutoregressive}; robot actions are high-dimensional vectors whose components can be determined in parallel~\citep{liu2024bidirectional, kim2025fine, song2025accelerating}. 
    Forcing both through a single autoregressive decoder creates a fundamental conflict: our ablation shows that naively adding CoT to an AR action decoder degrades performance from 85.5\% to 81.3\% while incurring $4\times$ latency (Table~\ref{CoT hybrid compare}).
    
    \textbf{(2)~Condition~2: Causal Alignment.} CoT must be causally linked to task success; otherwise, supervised fine-tuning (SFT) produces reasoning that merely imitates the style of expert annotations without influencing action selection. 
    The critical evidence: under OOD dynamics, our SFT-only model suffers a 32.0\,pp performance drop---nearly identical to the 31.6\,pp drop of the reasoning-free $\pi_0$\texttt{-FAST} baseline (Table~\ref{tab:libero_ood_mask}). This means SFT-learned CoT does not participate in adaptation at all. 
    Only after outcome-based reinforcement learning does the drop shrink to 24.4\,pp, and masking CoT at inference widens it back to 27.7\,pp---indicating that RL alignment makes the CoT pathway action-relevant, turning CoT from a passive narrative into a functional planning signal.

    Guided by these two conditions, we design \textbf{DeepThinkVLA}, a system in which every architectural and training choice is a direct consequence of satisfying Decoding Alignment and Causal Alignment.
    For Condition~1, we introduce a hybrid-attention decoder that generates CoT with causal attention and then switches to bidirectional attention for parallel action decoding, resolving the modality mismatch and reducing inference latency.
    For Condition~2, we employ a two-stage pipeline~\citep{zhang2024grape, kim2025robot, li2025simplevla}: SFT cold-starts the model with foundational reasoning on a synthetically constructed embodied CoT dataset, followed by outcome-based RL with sparse task-success rewards that causally align the full reasoning--action chain.
    The contribution is not that hybrid decoding, GRPO, or the generic SFT-then-RL recipe is novel in isolation~\citep{lu2025vla, song2026maniplvm}, but that identifying and jointly satisfying both conditions is what makes CoT work in VLA.

    Built on the public $\pi_{0}$\texttt{-FAST} weights~\citep{pertsch2025fast}, DeepThinkVLA achieves 97.0\% average success on LIBERO, setting a new state of the art.
    On the LIBERO-Plus robustness benchmark, it attains 79.0\% overall, exceeding $\pi_0$\texttt{-FAST} by 17.4 points.
    On the high-fidelity RoboTwin~2.0 benchmark, it reaches 59.3\% success, surpassing the strongest baseline by 21.7 points with particularly strong long-horizon performance.
    Controlled OOD experiments indicate that the gains stem from functional reasoning rather than memorized trajectories.

    In summary, our key contributions are: \textbf{(1)}~We identify and empirically validate two necessary conditions---Decoding Alignment and Causal Alignment---for Chain-of-Thought reasoning to be effective in VLA models, and demonstrate that violating either condition makes CoT harmful or decorative. \textbf{(2)}~We propose DeepThinkVLA, a principled system that jointly satisfies both conditions via a hybrid-attention decoder and a two-stage SFT-then-RL training pipeline. \textbf{(3)}~We achieve state-of-the-art results on three benchmarks (LIBERO, LIBERO-Plus, RoboTwin~2.0), with controlled ablations confirming that each condition is necessary for the observed gains. \textbf{(4)}~Finally, real-robot experiments provide preliminary evidence for the physical applicability of our CoT data construction and hybrid architecture.

	\section{Related Work}
    \textbf{Vision--Language--Action Models.}
    Recent progress in VLAs has focused on architectural innovations within the reactive perception-to-action paradigm. Early work such as RT-2~\citep{zitkovich2023rt} popularized this direction, leading to models that adopt diverse VLM backbones~\citep{black2024pi_0, bjorck2025gr00t, hung2025nora, pertsch2025fast} and train on large-scale robotic datasets~\citep{walke2023bridgedata, fang2023rh20t, o2024open, khazatsky2024droid, wu2024robomind}.
    Variants include diffusion-based decoders~\citep{black2024pi_0, liu2024rdt}, block-parallel decoders with bidirectional attention~\citep{liu2024bidirectional, kim2025fine, song2025accelerating}, and hierarchical structures separating planning from execution~\citep{belkhale2024rt, cui2025openhelix, team2025robobrain, bu2025agibot, intelligence2504pi0}.
    Across these designs, however, the action-generating policy remains reactive---mapping observations and goals directly to actions without explicit deliberation~\citep{sapkota2025vision}.
    None of these works addresses the question of how to effectively integrate explicit reasoning into the action-generation loop, which is the focus of our investigation.

    \textbf{Embodied Reasoning via Supervised Fine-Tuning.}
    To move beyond purely reactive mapping, several recent studies endow VLAs with embodied reasoning through SFT on CoT-augmented data~\citep{zawalski2024robotic, lin2025onetwovla, tan2025reason, chen2025training}. This line of work typically constructs CoT annotations for existing embodied datasets~\citep{o2024open}, often leveraging stronger cloud-based models~\citep{team2023gemini}, and then fine-tunes open-source VLAs such as OpenVLA~\citep{kim2024openvla}. While providing a first step toward ``think before acting,'' these methods face two persistent challenges: limited availability of high-quality CoT-annotated embodied data~\citep{wang2024all, xu2024survey, zhong2025survey}, and a tendency for SFT to produce shallow memorization of reasoning traces without strong alignment to actions~\citep{zhao2025embodied, liu2025spatialcot}.
    In the broader vision-language domain, SFT can induce pseudo reasoning paths in R1-like LVLMs while RL fosters more adaptive reasoning behavior~\citep{chen2025sft}, and carefully curated SFT data can substantially close multimodal reasoning gaps~\citep{lin2026mmfinereason}---indicating that SFT effectiveness is tightly coupled to data quality.
    Our Condition~2 targets a complementary, VLA-specific failure mode: unlike these settings, where the reasoning trace and final answer share a single textual stream, a VLA must causally link natural-language CoT to heterogeneous action outputs, so SFT alone cannot guarantee that the generated text influences action selection at any fixed data quality.
    Beyond SFT, recent RL-based efforts either optimize reasoning-free VLA backbones with outcome rewards~\citep{lu2025vla} or train LVLMs with verifiable rewards for affordance perception and image-space trajectory prediction without closed-loop action generation~\citep{song2026maniplvm}; our contribution is accordingly not the generic SFT-then-RL recipe, but the alignment of explicit reasoning with VLA action generation.
    Crucially, no prior work has systematically investigated under what conditions CoT reasoning transitions from decorative text to a functional decision-making signal. Our work fills this gap by identifying and validating two necessary conditions for effective CoT in VLA models.

	\section{DeepThinkVLA}
	\label{sec3}
    Guided by these two necessary conditions, we design DeepThinkVLA so that every architectural and training choice directly serves Decoding Alignment or Causal Alignment. We first formalize the ``think before acting'' paradigm as a probabilistic decomposition (Section~\ref{sec3-1}). We then present the hybrid-attention decoder, the natural architectural response to Condition~1 (Section~\ref{sec3-2}). Finally, we detail the two-stage training pipeline---SFT followed by outcome-based RL---which is the systematic implementation of Condition~2 (Section~\ref{sec3-3}).

	\subsection{Problem Formulation}
	\label{sec3-1}
    \begin{figure}[t!]
		\centering
		\includegraphics[width=0.95\columnwidth]{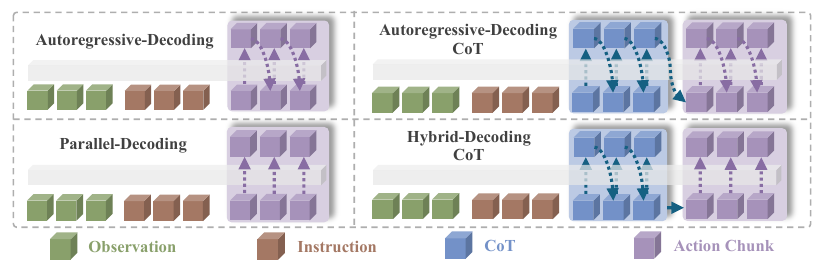}
		\caption{\textbf{Comparison of VLA architectures.}
        DeepThinkVLA introduces a hybrid design that pairs autoregressive CoT reasoning with parallel action decoding.}
		\label{fig:vla_architectures}
	\end{figure}
	Standard VLA policies learn a direct mapping from visual observations ($V$) and language instructions ($L$) to a sequence of actions ($A$).
    Instead, our work adopts the principle of ``think before acting''.
    We implement this principle by introducing a latent reasoning variable, the Chain-of-Thought ($R$). This approach decomposes the problem.
    Rather than modeling the direct policy, we model the joint probability of reasoning and then acting:
	\begin{equation}
		P(A, R \vert V, L) = P(A \vert V, L, R) \, P(R \vert V, L).
        \label{eq:joint_prob}
	\end{equation}

    The advantage of this decomposition is twofold.
    First, learning $P(R \vert V, L)$ is highly efficient. Most VLAs are built upon large VLM backbones, which already contain rich semantic and reasoning knowledge.
    Fine-tuning such models on a relatively small set of synthetically generated embodied CoT data is often sufficient to adapt their reasoning capability to the robotics domain.

    Second, learning $P(A \vert V, L, R)$ becomes significantly simpler than directly modeling $P(A\vert V, L)$.
    The CoT serves as an explicit, step-by-step plan that disambiguates high-level instructions $L$, turning the ill-posed, one-to-many mapping into a constrained and well-specified mapping from a reasoning step to its corresponding motor action.
    This principled factorization also enables emergent self-correction behaviors, as illustrated in Figure~\ref{cot-self-correct}.

    \subsection{Hybrid Architecture: Satisfying Decoding Alignment}
    \label{sec3-2}
    \begin{figure}[t!]
		\centering
		\includegraphics[width=0.95\columnwidth]{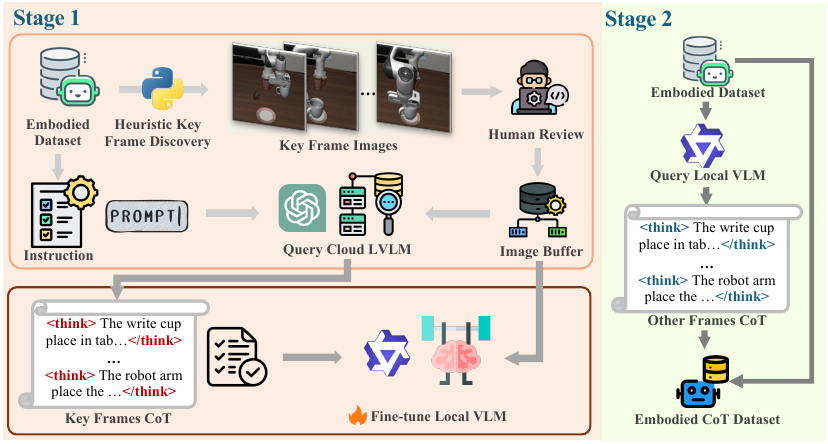}
		\caption{\textbf{Two-stage pipeline for embodied CoT dataset construction.}
			Stage~1 extracts keyframes via gripper state changes and queries a cloud LVLM;
			Stage~2 fine-tunes a local VLM on the keyframe CoT to annotate intermediate frames.}
		\label{data_construct}
	\end{figure}
    Condition~1 (Decoding Alignment) requires that CoT and actions be generated with mechanisms suited to their respective modalities. To implement the factorized policy from Eq.~\ref{eq:joint_prob} under this constraint, we design a hybrid architecture that aligns the decoding mechanism with the intrinsic properties of each modality. The core of our design is a dynamic attention mode within a single decoder: For CoT Generation ($P(R\vert V,L))$: The decoder employs standard autoregressive causal attention. This respects the sequential nature of language, where each reasoning token is generated based on its predecessors. For Action Generation ($P(A\vert V,L,R)$): After generating the CoT, the attention mechanism switches to bidirectional (non-causal) attention. This allows the model to process the entire action specification jointly and decode the action vector in parallel, acknowledging that different dimensions of a motor command (e.g., translation, rotation) are often determined concurrently.

    Beyond resolving this core modality mismatch, the parallel decoding of actions yields a critical practical advantage: a significant reduction in inference latency. This speedup is the key enabler for our subsequent training stage. While standard autoregressive models are prohibitively slow for the massive number of rollouts required by on-policy RL, our architecture's high-throughput action generation makes large-scale online fine-tuning computationally tractable. Further architectural details are summarized in Figure~\ref{fig:vla_architectures}; a token-level specification of the hybrid decoder---action-slot representation, embedding zeroing, and the hybrid attention mask---is provided in Appendix~\ref{appendix:hybrid_decoder}.

	\subsection{RL Training Pipeline: Satisfying Causal Alignment}
    \label{sec3-3}
    Condition~2 (Causal Alignment) demands that CoT be causally linked to task success rather than merely imitating expert annotation style. With our efficient architecture in place, we train DeepThinkVLA using a two-stage pipeline designed to first instill foundational reasoning and then establish the causal link through outcome-based optimization.

	\textbf{SFT Cold-Start for Foundational Reasoning. }
    SFT cold-start is designed to equip the model with a foundational CoT reasoning capability. This phase requires a specific supervision format comprising complete $(V, L, R, A)$ sequences. However, most existing large-scale embodied datasets lack explicit CoT annotations and instead provide only $(V, L, A)$ tuples. To address this critical data gap, we developed a scalable, two-stage data augmentation pipeline that generates high-quality CoT annotations, as illustrated in Figure~\ref{data_construct}.

    Our pipeline is optimized for both annotation quality and cost-efficiency.
    In stage 1, we identify semantically significant keyframes within each trajectory by detecting changes in the gripper state, which often indicate subtask boundaries. For these keyframes, CoT annotations are obtained by querying a powerful, general-purpose cloud-based Vision-Language Model (VLM) (see prompt in Appendix Fig.~\ref{data_construct_prompt}).
    In stage 2, to efficiently annotate the numerous intermediate frames, we fine-tune a smaller, locally-deployed VLM on the high-quality keyframe annotations obtained in Stage 1. This specialized model then automatically generates CoT annotations for the transitional frames. To ensure data fidelity, we apply schema checks to filter malformed outputs and enforce temporal consistency, resulting in a uniform embodied CoT dataset suitable for SFT; text-level quality audits of the resulting annotations are reported in Appendix~\ref{appendix:cot_audit}.

	\textbf{Learning Reasoning and Action via RL. }
    \begin{figure}[t!]
		\centering
		\includegraphics[width=0.99\textwidth]{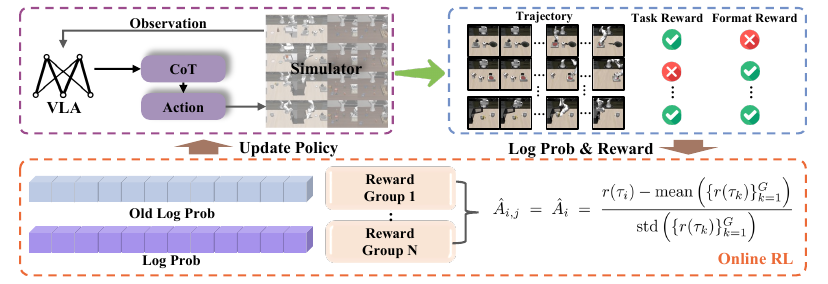}
		\caption{\textbf{RL stage with grouped credit assignment.}
        Trajectories are collected via simulator rollouts; rewards are group-normalized into token-level advantages and optimized with a clipped surrogate objective and KL regularization.}
		\label{RL_framework}
	\end{figure}
    While SFT provides a strong imitation learning foundation, it cannot adapt to novel scenarios or optimize beyond the sub-optimal trajectories present in the static dataset. To overcome these limitations and truly learn a policy that maximizes task success, we introduce a second training stage based on online Reinforcement Learning. We use an outcome-based reward to jointly optimize the entire reasoning-action sequence towards the singular goal of task completion (Figure~\ref{RL_framework}).

    We cast the online RL stage as a policy optimization problem: starting from the initial SFT policy, we aim to maximize the expected outcome-based reward through interactive learning. We adopt an on-policy policy gradient algorithm with a clipped surrogate objective akin to PPO. The VLA collects trajectories from the environment, computes advantages, and updates its policy accordingly.

    Formally, during the RL rollout, the state at each step $t$ is defined as
    $s_t = [o_t^{\text{vis}}, \ell_{\text{task}}]$,
    where $o_t^{\text{vis}}$ denotes the visual observation and $\ell_{\text{task}}$ is the task instruction.
    Given the state input, the VLA outputs
    $\mathcal{A}_t = [a_t^{\text{cot}}, a_t^{\text{robot}}]$,
    where the reasoning tokens $a_t^{\text{cot}}$ are generated autoregressively, and the action tokens
    $a_t^{\text{robot}} \in \mathbb{R}^{h \times d}$ are decoded in parallel.
    Here, $h$ denotes the action chunk size and $d$ corresponds to the robot's control dimension (e.g., $d=7$ for a 6-DoF manipulator plus gripper control).
    \begin{table*}[t!]
    	\centering
    	\scriptsize
    	\setlength{\tabcolsep}{5pt}
    	\caption{\textbf{Performance on the LIBERO simulation benchmark} (success rate \%, 50 trials per task).
        All top baselines use wrist cameras; DeepThinkVLA uses vision-language only. \textbf{Bold} = best per suite.
        }
        \label{table1}
    	\resizebox{0.99\linewidth}{!}{
    		\begin{tabular}{l|l *{5}{c}}
    			\toprule
                \textbf{Category}
                & \textbf{Method}
                & \textbf{Object}
                & \textbf{Spatial}
                & \textbf{Goal}
                & \textbf{Long}
                & \textbf{Average} \\
    			\midrule
    			\multirow{9}*{\textbf{AR-Decoding}}
    			    & TraceVLA~\citep{zheng2024tracevla}  & 85.2 & 84.6 & 75.1 & 54.1 & 74.8 \\
    			    & Octo~\citep{team2024octo}           & 85.7 & 78.9 & 84.6 & 51.1 & 75.1 \\
    			    & OpenVLA~\citep{kim2024openvla}      & 88.4 & 84.7 & 79.2 & 53.7 & 76.5 \\
    			    & SpatialVLA~\citep{qu2025spatialvla} & 89.9 & 88.2 & 78.6 & 55.5 & 78.1 \\
                    & GRAPE~\citep{zhang2024grape}        & 91.2 & 87.6 & 82.2 & 55.8 & 79.2 \\
    			    & NORA~\citep{hung2025nora}           & 95.4 & 92.2 & 89.4 & 74.6 & 87.9 \\
                    & VLA-RL~\citep{lu2025vla}            & 91.8 & 90.2 & 82.2 & 59.8 & 81.0 \\
    			    & \(\pi_{0}\)-FAST~\citep{pertsch2025fast} & 96.8 & 96.4 & 88.6 & 60.2 & 85.5 \\
    			    & UniVLA~\citep{bu2025univla}         & 96.8 & 96.5 & 95.6 & 92.0 & 95.2 \\
    			\midrule
    			\multirow{2}{*}{\textbf{Diffusion}}
    			    & Diffusion Policy~\citep{chi2023diffusion} & 92.5 & 78.3 & 68.3 & 50.5 & 72.4 \\
                    & \(\pi_{0}\)~\citep{black2024pi_0}   & 98.8 & \textbf{96.8} & 95.8 & 85.2 & 94.2 \\
    			\midrule
                \multirow{2}{*}{\textbf{Parallel-Decoding}}
                & CoT-VLA-7B~\citep{zhao2025cot}      & 91.6 & 87.5 & 87.6 & 69.0 & 81.1 \\
                & OpenVLA--OFT~\citep{kim2025fine} & 92.7 & 91.3 & 90.5 & 86.5 & 90.3 \\
                \midrule
    			\multirow{2}{*}{\textbf{Hybrid-Decoding}}
    			    & DeepThinkVLA (Qwen3-VL) & 98.6 & 92.6 & 96.2 & 92.0 & 94.9 \\
    			    & DeepThinkVLA ($\pi_0$-FAST) & \textbf{99.0} & 96.6 & \textbf{96.4} & \textbf{96.2} & \textbf{97.0} \\
    			\bottomrule
    		\end{tabular}
        	}
    \end{table*}

    Let $\pi_\theta$ denote the current policy. A trajectory $\tau$ sampled from the old policy $\pi_{\theta_{\text{old}}}$ is defined as $\tau = [(s_0, \mathcal{A}_0), (s_1, \mathcal{A}_1), \dots, (s_T, \mathcal{A}_T)]$.
    The reward function $\mathcal{R}(\tau)$ is sparse, awarded only at the end of the trajectory based on a verifiable task completion signal $\mathcal{I}_{\text{success}}$. No intermediate reward is given for the semantics of the reasoning trace. A small format-regularization reward $\mathcal{I}_{\text{format}}$ is added to prevent stylistic drift.  Hence, the reward function is defined as
    \begin{equation}
        \begin{aligned}
            \mathcal{R}(\tau)
            = \alpha_{s} & \cdot \mathcal{I}_{\text{success}} + \alpha_{f} \cdot \mathcal{I}_{\text{format}}, \\
            \mathcal{I}_{\text{success}} =
            \begin{cases}
                1, & \text{if task success}, \\
                0, & \text{otherwise},
            \end{cases}
            &\quad\mathcal{I}_{\text{format}} =
            \begin{cases}
                1, & \text{if CoT format correct}, \\
                0, & \text{otherwise}.
            \end{cases}
        \end{aligned}
        \label{eq:reward_function}
    \end{equation}
    where $\alpha_{s}$ and $\alpha_{f}$ are weighting coefficients.
    Then the token-level clipped surrogate objective is:
	\begin{equation}
		\label{eq:ppo_generic}
		\mathcal{J}(\theta)= \mathbb{E}_{\tau \sim \pi_{\theta_{\text{old}}}}
			\Bigg[ \sum_{j=1}^{N} \min\!\Big(\omega_j(\theta)\,\hat{A}_j,\;\operatorname{clip}\!\big(\omega_j(\theta), 1-\epsilon, 1+\epsilon\big)\,\hat{A}_j
			\Big) \Bigg],
	\end{equation}
	where $N = \lvert \mathcal{A}_t \rvert \times T$ denotes the total number of tokens in a trajectory $\tau$. $\omega_j(\theta) = \frac{\pi_\theta(a_j\vert s_t, a_{<j})}{\pi_{\theta_{\text{old}}}(a_j\vert s_t, a_{<j})}$ and $\hat{A}_j$ are the importance ratio and the advantage for token $a_j \in [\mathcal{A}_0, \dots, \mathcal{A}_T]$ within trajectory $\tau$, respectively.

    To propagate the sparse, outcome-based reward $\mathcal{R}(\tau)$ to each token prediction, we adopt the credit assignment strategy from GRPO~\citep{shao2024deepseekmath}. A group of $G$ trajectories is collected for each task prompt and their rewards are standardized to compute a shared advantage value for all tokens within a given trajectory. For token $j$ in trajectory $i$, the advantage is:
	\begin{equation}
		\hat{A}_{i,j} \;=\; \frac{\mathcal{R}(\tau_i) - \operatorname{mean}\big(\{\mathcal{R}(\tau_k)\}_{k=1}^{G}\big)}{\operatorname{std}\big(\{\mathcal{R}(\tau_k)\}_{k=1}^{G}\big)}.
	\end{equation}
	This relative credit assignment encourages the model to prefer reasoning and action sequences that lead to better-than-average outcomes, effectively selecting for more functional thought processes.

    Combining the clipped surrogate objective with the GRPO-style advantage and a KL-divergence penalty to the original SFT policy $\pi_{\text{ref}}$, which prevents catastrophic forgetting, our final objective is:
	\begin{equation}
		\begin{aligned}
			\mathcal{J}_{\text{final}}(\theta)&= \mathbb{E}_{\substack{s \sim \text{env},\\ \{\tau_{i}\}_{i=1}^G \sim \pi_{\theta_{\text{old}}}}}
			\Bigg[
			\frac{1}{G}\sum_{i=1}^G \frac{1}{N}\sum_{j=1}^{N}
			\min\!\Big(
			\omega_{i,j}(\theta)\,\hat{A}_{i,j},\; \operatorname{clip}\!\big(\omega_{i,j}(\theta), 1-\epsilon, 1+\epsilon\big)\,\hat{A}_{i,j}
			\Big) \\
            &- \beta\,\mathrm{KL}\!\big(\pi_\theta(\cdot\mid s)\,\|\allowbreak\,\pi_{\text{ref}}(\cdot\mid s)\big)
			\Bigg],
		\end{aligned}
		\label{GRPO_objective_final}
	\end{equation}
    where $\omega_{i,j}(\theta)$ is the importance ratio for token $a_j \in [\mathcal{A}_0, \dots, \mathcal{A}_T]$ in trajectory $i$.
	By maximizing $\mathcal{J}_{\text{final}}$, the VLA simultaneously refines its reasoning and action abilities, with both aligned toward the singular goal of maximizing the final task success rate.

    \begin{table*}[tb]
        \centering
        \scriptsize
        \setlength{\tabcolsep}{7pt}
        \renewcommand{\arraystretch}{1}
        \definecolor{rowblue}{RGB}{220,230,255}
        \definecolor{gain}{RGB}{0,140,70}
        \caption{\textbf{Performance on RoboTwin 2.0} (success rate \%).
        DeepThinkVLA achieves \textbf{59.3\%} overall, exceeding $\pi_0$-FAST by \textbf{+21.7} points, with the largest gains on long-horizon tasks.
        }
        \label{tab:robotwin}
        \resizebox{\linewidth}{!}{
        % \rowcolors*{1}{yellow!100}{yellow!100}
        \begin{tabular}{lccccc}
        \toprule
        \multicolumn{6}{c}{\textbf{Short Horizon Tasks (100--130 Steps)}} \\
        \midrule
        \textbf{Model} & \textbf{Lift Pot} & \textbf{Beat Hammer Block} & \textbf{Pick Dual Bottles} & \textbf{Place Phone Stand} & \textbf{Avg} \\
        \midrule
        $\pi_0$~\citep{black2024pi_0}      & 51.0 & 59.0 & 50.0 & 22.0 & 45.5 \\
        RDT~\citep{liu2024rdt}          & 45.0 & 22.0 & 18.0 & 13.0 & 24.5 \\
        OpenVLA-OFT~\citep{kim2025fine}  & 10.1 & 28.1 & 29.7 & 17.1 & 21.3 \\
        \midrule
        $\pi_{0}$\texttt{-FAST}~\citep{pertsch2025fast}  & 30.0 & 38.0 & 25.0 & 16.0 & 27.3 \\
        \rowcolor{rowblue}
        \textbf{DeepThinkVLA} & \textbf{62.0} & \textbf{73.0} & \textbf{61.0} & \textbf{24.0} & \textbf{55.0} \\
        $\Delta$     & \textcolor{gain}{$\uparrow$32.0} & \textcolor{gain}{$\uparrow$35.0} &
                       \textcolor{gain}{$\uparrow$36.0} & \textcolor{gain}{$\uparrow$8.0} &
                       \textcolor{gain}{$\uparrow$27.8} \\
        \midrule
        \multicolumn{6}{c}{\textbf{Medium Horizon Tasks (150--230 Steps)}} \\
        \midrule
        \textbf{Model} & \textbf{Move Can Pot} & \textbf{Place A2B Left}
                       & \textbf{Place Empty Cup} & \textbf{Handover Mic} & \textbf{Avg} \\
        \midrule
        $\pi_0$~\citep{black2024pi_0}      & 41.0 & 38.0 & 60.0 & 96.0 & 58.8 \\
        RDT~\citep{liu2024rdt}           & 33.0 & 21.0 & 42.0 & 95.0 & 47.8 \\
        OpenVLA-OFT~\citep{kim2025fine}  & 28.1 & 37.5 & 77.3 & 45.3 & 47.1 \\
        \midrule
        $\pi_{0}$\texttt{-FAST}~\citep{pertsch2025fast}  & 34.0 & 36.0 & 54.0 & 83.0 & 51.8 \\
        \rowcolor{rowblue}
        \textbf{DeepThinkVLA} & \textbf{52.0} & \textbf{38.0} & \textbf{83.0} & \textbf{88.0} & \textbf{65.3} \\
        $\Delta$     & \textcolor{gain}{$\uparrow$18.0} & \textcolor{gain}{$\uparrow$2.0} &
                       \textcolor{gain}{$\uparrow$29.0} & \textcolor{gain}{$\uparrow$5.0} &
                       \textcolor{gain}{$\uparrow$13.5} \\
        \midrule
        \multicolumn{6}{c}{\textbf{Long (280--320 Steps) \& Extra Long Horizon Tasks (450--650 Steps)}} \\
        \midrule
        \textbf{Model} & \textbf{Handover Block} & \textbf{Stack Bowls Two}
                       & \textbf{Blocks Rank Rgb} & \textbf{Put Bottles Dustbin} & \textbf{Avg} \\
        \midrule
        $\pi_0$~\citep{black2024pi_0}      & 39.0 & 53.0 & 45.0 & 36.0 & 43.3 \\
        RDT~\citep{liu2024rdt}          & 26.0 & 42.0 & 17.0 & 26.0 & 27.8 \\
        OpenVLA-OFT~\citep{kim2025fine}  & 33.1 & 40.6 & 70.2 & 42.2 & 46.5 \\
        \midrule
        $\pi_{0}$\texttt{-FAST}~\citep{pertsch2025fast}  & 32.0 & 48.0 & 28.0 & 27.0 & 33.8 \\
        \rowcolor{rowblue}
        \textbf{DeepThinkVLA} & \textbf{43.0} & \textbf{62.0} & \textbf{77.0} & \textbf{49.0} & \textbf{57.8} \\
        $\Delta$     & \textcolor{gain}{$\uparrow$11.0} & \textcolor{gain}{$\uparrow$14.0} &
                       \textcolor{gain}{$\uparrow$49.0} & \textcolor{gain}{$\uparrow$22.0} &
                       \textcolor{gain}{$\uparrow$24.0} \\
        \midrule
        \multicolumn{6}{l}{\textbf{Overall Avg}\quad
        RDT: 33.3 \quad $\pi_0$: 49.2 \quad OpenVLA-OFT: 38.3 \quad $\pi_{0}$\texttt{-FAST}: 37.6 \quad
        \textbf{DeepThinkVLA: 59.3} \quad \textcolor{gain}{$\uparrow$21.7}} \\
        \bottomrule
        \end{tabular}
        }
    \end{table*}

	\section{Experiments}
	\label{Experiments}
	\subsection{Experimental Setup}
    \textbf{Implementation Details.}
    DeepThinkVLA is initialized from the public \(\pi_{0}\)\texttt{-FAST} weights~\citep{pertsch2025fast}.
    We refactor the baseline policy with our hybrid-attention decoder (Sec.~\ref{sec3-2}), yielding a 2.9B parameter model.
    Training proceeds in two stages. First, an SFT cold-start is performed on our embodied CoT dataset. A hybrid attention mask supervises CoT tokens with causal attention and action tokens with bidirectional attention in a single forward pass, optimized by token-level cross-entropy loss.
    Second, we apply online RL to align generated CoT with action execution, using outcome-based rewards and GRPO-style grouped credit assignment.
    Additional training details, including dataset statistics, hyperparameters, and inference settings, are provided in Appendix~\ref{appendix:implementation}.

    \textbf{Benchmarks.}
    We evaluate DeepThinkVLA on three simulation benchmarks and a set of real-robot tasks to ensure a comprehensive assessment of foundational skills, long-horizon robustness, and physical transferability.
    LIBERO~\citep{liu2023libero}: A standard language-conditioned manipulation benchmark containing four suites (Object, Spatial, Goal, and Long). Evaluation is conducted under 50 randomized initial conditions for each task.

    LIBERO-Plus~\citep{fei2025libero}: A robustness benchmark that introduces controlled perturbations across seven dimensions (e.g., camera viewpoints, robot initial states, and language rewrites) and reports per-dimension success rates along with an overall score.
    RoboTwin 2.0~\citep{chen2025robotwin}: A high-fidelity digital twin benchmark featuring complex contact-rich manipulation and significantly longer task horizons. This is introduced to rigorously test the model's capability to maintain reasoning context over extended execution periods.
    Real-Robot Tasks: To assess the preliminary physical applicability of our CoT data construction and hybrid architecture, we additionally evaluate on three physical manipulation tasks (Stack Bowls Two, Handover Block, Blocks Rank RGB) using human-teleoperated demonstrations and our CoT annotation pipeline (details in Section~\ref{sec:real_robot}).
    Performance is measured by the average success rate. Additionally, to provide concrete examples of the model's reasoning process and error recovery behaviors (Case Studies), we include a detailed qualitative analysis in Appendix~\ref{sec:appendix_casestudy}.

    \textbf{Baselines.}
    We evaluate DeepThinkVLA against a wide spectrum of recent VLA systems, covering autoregressive SFT models, diffusion-based approaches, parallel-decoding methods and commercial baselines such as $\pi_{0}$ and $\pi_{0}$\texttt{-FAST}.
    A detailed description of all baselines is provided in Appendix~\ref{appendix:baselines}.

	\subsection{Main Results}
    \textbf{Results on LIBERO.}
    The main results on the LIBERO benchmark are shown in Table~\ref{table1}. DeepThinkVLA establishes a new state-of-the-art by achieving the highest average success rate (SR) of 97.0\%. Our model shows exceptional proficiency across all categories, particularly in Object (99.0\%) and long-horizon tasks (96.2\%), significantly outperforming leading diffusion models like $\pi_{0}$ (Object: 98.8\%, Long: 85.2\%) and autoregressive baselines like UniVLA (Object: 96.8\%, Long: 92.0\%).
    \begin{table*}[tb]
      \centering
      \caption{\textbf{Zero-shot robustness on LIBERO-Plus} (seven perturbation dimensions, no adaptation).
      OFT\_w = wrist-camera variant; OFT\_m = mixed-data variant. \textbf{Bold} = best per dimension.
      }
      \label{tab:perturb_robustness}
      \scriptsize
      \setlength{\tabcolsep}{3pt}
      \renewcommand{\arraystretch}{1.05}

      % colors
      \definecolor{sectiongray}{RGB}{235,235,235}
      \definecolor{rowblue}{RGB}{220,230,255}
      \definecolor{gain}{RGB}{0,140,70}

      \resizebox{0.99\linewidth}{!}{%
      \begin{tabular}{lcccccccc}     \toprule
        \textbf{Model} &
        \multicolumn{7}{c}{\textbf{Perturbation Dimensions}} &
        \textbf{Total} \\
        \cmidrule(lr){2-8}
        & \textbf{Camera} & \textbf{Robot-Init} & \textbf{Language} & \textbf{Light} &
          \textbf{Background} & \textbf{Noise} & \textbf{Layout} & \\
        \midrule

        $\pi_0$~\citep{black2024pi_0}             & 13.8 & 6.0  & 58.8 & 85.0 & 81.4 & 79.0 & 68.9 & 53.6 \\
        UniVLA~\citep{bu2025univla}          & 1.8  & \textbf{46.2} & 69.6 & 69.0 & 81.0 & 21.2 & 31.9 & 42.9 \\
        WorldVLA~\citep{cen2025worldvla}        & 0.1  & 27.9 & 41.6 & 43.7 & 17.1 & 10.9 & 38.0 & 25.0 \\
        OpenVLA~\citep{kim2024openvla}         & 0.8  & 3.5  & 23.0 & 8.1  & 34.8 & 15.2 & 28.5 & 15.6 \\
        OpenVLA-OFT~\citep{kim2025fine}     & 56.4 & 31.9 & 79.5 & 88.7 & 93.3 & 75.8 & 74.2 & 69.6 \\
        OpenVLA-OFT\_w  & 10.4 & 38.7 & 70.5 & 76.8 & \textbf{93.6} & 49.9 & 69.9 & 55.8 \\
        OpenVLA-OFT\_m  & 55.6 & 21.7 & 81.0 & 92.7 & 91.0 & 78.6 & 68.7 & 67.9 \\
        NORA-long~\citep{hung2025nora}       & 2.2  & 37.0 & 65.1 & 45.7 & 58.6 & 12.8 & 62.1 & 39.0 \\
        RIPT-VLA~\citep{tan2025interactive}        & 55.2 & 31.2 & 77.6 & 88.4 & 91.6 & 73.5 & 74.2 & 68.4 \\
        \midrule

        $\pi_0$-FAST~\citep{pertsch2025fast}        & 65.1 & 21.6 & 61.0 & 73.2 & 73.2 & 74.4 & 68.8 & 61.6 \\
        \midrule
        DeepThinkVLA (Qwen3-VL) & 63.7 & 38.8 & 81.5 & \textbf{94.7} & 92.4 & 89.8 & 77.8 & 77.0 \\
        \rowcolor{rowblue}
        \textbf{DeepThinkVLA ($\pi_0$-FAST)} & \textbf{88.5} & 40.5 & \textbf{84.5} & 90.0 & 75.3 & \textbf{94.4} & \textbf{79.9} & \textbf{79.0} \\
        \bottomrule
          \end{tabular}%
      }
    \end{table*}

    Notably, all competitive baselines in Table~\ref{table1} utilize wrist camera inputs, while DeepThinkVLA achieves this performance using a pure vision-language setup without proprioceptive state.

    \textbf{Results on RoboTwin 2.0.}
    On the high-fidelity RoboTwin 2.0 benchmark (Table~\ref{tab:robotwin}), DeepThinkVLA achieves 59.3\% average success, outperforming $\pi_{0}$\texttt{-FAST} (37.6\%) by +21.7 points. The gains are most pronounced on long and extra-long horizon tasks (57.8\% vs.\ 33.8\%), confirming that explicit reasoning helps maintain context over extended execution.

    \textbf{Results on LIBERO-Plus.}
    As shown in Table~\ref{tab:perturb_robustness}, DeepThinkVLA achieves 79.0\% overall on the seven-dimension robustness benchmark, outperforming $\pi_0$\texttt{-FAST} by +17.4 points. The gains are broad-based: Camera +23.4, Language +23.5, Noise +20.0, confirming generalization under substantial distribution shifts.

    \subsection{Real-Robot Experiments}
    \label{sec:real_robot}
    \begin{wraptable}{r}{0.45\linewidth}
        \vspace{-10pt}
        \centering
        \caption{\textbf{Real-robot results} (\%, 20 trials).}
        \label{tab:real_robot}
        \scriptsize
        \renewcommand{\arraystretch}{1.15}
        \setlength{\tabcolsep}{3pt}
        \begin{tabular}{c|ccc|c}
            \toprule
            \multirow{2}{*}{\textbf{Task}} & \textbf{Stack} & \textbf{Handover} & \textbf{Blocks} & \multirow{2}{*}{\textbf{Avg}} \\
            & \textbf{Bowls} & \textbf{Block} & \textbf{Rank RGB} & \\
            \midrule
            \textbf{Performance} & \textcolor{blue}{55\%} & \textcolor{blue}{45\%} & \textcolor{blue}{35\%} & \textcolor{blue}{45\%} \\
            \bottomrule
        \end{tabular}
        \vspace{-10pt}
    \end{wraptable}
    As a preliminary physical-applicability check, we deploy DeepThinkVLA on an AGILEX ALOHA bimanual platform for three manipulation tasks (Stack Bowls Two, Handover Block, Blocks Rank RGB): teleoperated demonstrations are annotated with the same two-stage CoT pipeline (Section~\ref{sec3-3}), and the model is fine-tuned via SFT only---performing online RL on hardware would require many unsafe trial-and-error rollouts, so claims about RL-based causal alignment remain grounded in simulation.
    As shown in Table~\ref{tab:real_robot}, DeepThinkVLA attains 45\% average success over 20 trials per task.
    This figure is best compared against the harder, contact-rich RoboTwin~2.0 simulation results on the corresponding task categories (62.0\% / 43.0\% / 77.0\%; Table~\ref{tab:robotwin}) rather than near-saturated LIBERO: the physical results are lower but remain within the same order of difficulty, showing that the pipeline can be instantiated on hardware and achieve nontrivial closed-loop execution under real camera, calibration, contact, and actuation noise.
    This evidence supports the preliminary physical applicability of the CoT data construction and hybrid action decoder; it does not validate the full SFT-then-RL pipeline in the physical domain. Task definitions and further discussion are provided in Appendix~\ref{appendix:real_robot_details}.

    \subsection{Ablation Studies}
    \label{sec:ablations}

    \textbf{Backbone Generality.}
    To test whether our two conditions are architecture-specific or general principles, we apply the same hybrid-decoding design and CoT annotation pipeline to Qwen3-VL~\citep{bai2025qwen3}---a VLM without any embodied pretraining. As shown in Tables~\ref{table1} and~\ref{tab:perturb_robustness}, DeepThinkVLA (Qwen3-VL) achieves 94.9\% on LIBERO and 77.0\% on LIBERO-Plus zero-shot, both substantially above all baselines except the $\pi_0$-FAST-based variant. This confirms that the two conditions---not the choice of pretrained backbone---are the primary drivers of effective embodied reasoning. A model without any robotic pretraining can still acquire strong manipulation and robustness through proper decoding alignment and CoT construction. The gains are likewise not driven by sensor modalities: even without a wrist camera, DeepThinkVLA reaches 86.0\% on LIBERO, already surpassing the $\pi_0$-FAST baseline (85.5\%; Appendix~\ref{appendix:robustness}).

    \begin{table}[t!]
        \centering
        \caption{\textbf{Same-architecture w/o-CoT control on LIBERO} (success rate \%). The action decoder and RL procedure are identical across rows; only the CoT pathway differs. RL alone yields +1.1\,pp over the w/o-CoT SFT model, while retaining CoT under the identical RL procedure adds a further +3.2\,pp, with the largest gains on Goal (+4.6\,pp) and Long (+6.0\,pp).}
        \label{tab:no_cot_control}
        \small
        \setlength{\tabcolsep}{6pt}
        \renewcommand{\arraystretch}{1.05}
        \begin{tabular}{lccccc}
        \toprule
        \textbf{Method} & \textbf{Object} & \textbf{Spatial} & \textbf{Goal} & \textbf{Long} & \textbf{Avg.} \\
        \midrule
        DeepThinkVLA (w/o CoT, SFT-Only)    & 97.2 & 96.4 & 89.6 & 87.8 & 92.7 \\
        DeepThinkVLA (w/o CoT, RL-Aligned)  & 97.2 & 96.2 & 91.8 & 90.2 & 93.8 \\
        DeepThinkVLA (Full CoT, RL-Aligned) & \textbf{99.0} & \textbf{96.6} & \textbf{96.4} & \textbf{96.2} & \textbf{97.0} \\
        \bottomrule
        \end{tabular}
    \end{table}

    \textbf{Decoding Alignment.}
    To directly validate Condition~1, we apply identical embodied CoT supervision to the fully autoregressive $\pi_0$-FAST model: naively decoding CoT and actions through a single AR stream drops the LIBERO average from 85.5\% to 81.3\% while incurring $4\times$ inference latency, whereas the hybrid decoder reaches 96.8\% with parallel action decoding (Table~\ref{CoT hybrid compare} in Appendix~\ref{sec:appendix_arch})---a +15.5\,pp swing attributable solely to matching each modality with its appropriate decoding mechanism.
    The parallel action pass also keeps inference at $0.175\times$--$1.4\times$ of the baseline (vs.\ $4.0\times$ for AR-CoT), which is what makes the massive on-policy rollouts of the RL stage computationally tractable.

    \textbf{Causal Alignment via RL.}
    Applying RL improves LIBERO-Long from 94.2\% to 96.2\% (Figure~\ref{fig:rl_libero_long} in Appendix~\ref{appendix:robustness}), with a more pronounced +6.8\% gain on the harder RoboTwin 2.0 (Appendix~\ref{sec:appendix_robotwin_rl}). The controlled OOD experiment (Table~\ref{tab:libero_ood_mask} in Appendix~\ref{appendix:robustness}) provides the strongest evidence: SFT-only suffers a 32.0\,pp drop under Joint-Limit dynamics---nearly identical to the reasoning-free $\pi_0$-FAST (31.6\,pp)---while RL-aligned DeepThinkVLA limits the drop to 24.4\,pp. Masking CoT widens it to 27.7\,pp, indicating that RL alignment makes the CoT pathway action-relevant rather than decorative.
    A same-architecture control that removes CoT throughout SFT, RL, and inference (Table~\ref{tab:no_cot_control}) further separates the contribution of the CoT pathway from that of the decoder and RL stage: RL without CoT reaches 93.8\% on LIBERO, while retaining CoT under the identical RL procedure reaches 97.0\% (+3.2\,pp, with +4.6\,pp on Goal and +6.0\,pp on Long).
    We therefore state Condition~2 precisely: Causal Alignment is required for CoT to remain functional in \emph{action-execution-sensitive} OOD settings; under perception/instruction-side shifts, RL alignment preserves the broad zero-shot robustness of the SFT model (79.1\% vs.\ 79.0\% overall on LIBERO-Plus; Table~\ref{tab:libero_plus_rl} in Appendix~\ref{appendix:robustness}).
    Further architectural ablations (Hybrid vs.\ AR) and CoT semantic analyses are provided in Appendix~\ref{sec:appendix_arch}, and text-level quality audits of demonstration and deployment-time CoTs in Appendix~\ref{appendix:cot_audit}.

	\section{Conclusions}
    We identified two necessary conditions for CoT reasoning to be effective in VLA models---\textbf{Decoding Alignment} and \textbf{Causal Alignment}---and showed that violating either renders CoT harmful or decorative: misaligned decoding actively degrades performance, while Causal Alignment is required for CoT to remain functional in action-execution-sensitive OOD settings. DeepThinkVLA jointly satisfies both conditions via a hybrid-attention decoder and SFT-then-RL training, achieving state-of-the-art results on LIBERO, LIBERO-Plus, and RoboTwin~2.0. Backbone ablations on Qwen3-VL confirm that the conditions, not the pretrained weights, drive the gains. Real-robot experiments provide preliminary evidence for the physical applicability of the CoT data construction and hybrid architecture. The main limitations are that the physical evidence remains SFT-only and small-scale, and that our causal analysis is functional rather than mechanistic; safely extending online RL to hardware and probing the internal mechanism are important next steps. We believe this two-condition framework offers a useful lens for future embodied reasoning research; as one such direction, Appendix~\ref{appendix:spatial_reasoning} presents a preliminary exploration of augmenting language CoT with explicit spatial reasoning.

% \section*{Acknowledgments}
% Use unnumbered first level headings for the acknowledgments. All

\newpage
\section*{Acknowledgement}
We sincerely thank all the anonymous reviewers
and (S)ACs for their constructive comments and
helpful suggestions. This work was supported by Beijing Major Science and Technology Project under Contract no.Z251100008125046, the National Natural Science Foundation of China under Grant 52188102 and the Hubei Province Key Research and Development Project (No. 2025BEB008)
\section*{Ethics Statement}
This work develops methods to improve robotic manipulation through explicit reasoning in controlled environments. All real-robot experiments are safely conducted on an AGILEX ALOHA platform under direct human supervision using benign objects. Our training relies strictly on public benchmarks (LIBERO, RoboTwin~2.0) and consensual laboratory demonstrations, involving no sensitive or personal data. While the reinforcement learning stage incurs computational costs across an 8-GPU cluster, we mitigate this environmental impact through an efficient hybrid-decoding architecture and parallelized rendering. Ultimately, by rigorously validating the functional role of reasoning in Vision--Language--Action models, we aim to contribute to the development of safer, more transparent, and interpretable autonomous systems.

\bibliography{colm2026_conference}

@article{liu2025aligning,
	title={Aligning cyber space with physical world: A comprehensive survey on embodied ai},
	author={Liu, Yang and Chen, Weixing and Bai, Yongjie and Liang, Xiaodan and Li, Guanbin and Gao, Wen and Lin, Liang},
	journal={IEEE/ASME Transactions on Mechatronics},
	year={2025},
	publisher={IEEE}
}

@article{wang2024all,
	title={All robots in one: A new standard and unified dataset for versatile, general-purpose embodied agents},
	author={Wang, Zhiqiang and Zheng, Hao and Nie, Yunshuang and Xu, Wenjun and Wang, Qingwei and Ye, Hua and Li, Zhe and Zhang, Kaidong and Cheng, Xuewen and Dong, Wanxi and others},
	journal={arXiv preprint arXiv:2408.10899},
	year={2024}
}

@article{zhong2025survey,
	title={A Survey on Vision-Language-Action Models: An Action Tokenization Perspective},
	author={Zhong, Yifan and Bai, Fengshuo and Cai, Shaofei and Huang, Xuchuan and Chen, Zhang and Zhang, Xiaowei and Wang, Yuanfei and Guo, Shaoyang and Guan, Tianrui and Lui, Ka Nam and others},
	journal={arXiv preprint arXiv:2507.01925},
	year={2025}
}

@inproceedings{yang2024robot,
	title={Robot fine-tuning made easy: Pre-training rewards and policies for autonomous real-world reinforcement learning},
	author={Yang, Jingyun and Mark, Max Sobol and Vu, Brandon and Sharma, Archit and Bohg, Jeannette and Finn, Chelsea},
	booktitle={IEEE International Conference on Robotics and Automation},
	pages={4804--4811},
	year={2024},
	organization={IEEE}
}

@inproceedings{
	kim2024openvla,
	title={Open{VLA}: An Open-Source Vision-Language-Action Model},
	author={Moo Jin Kim and Karl Pertsch and Siddharth Karamcheti and Ted Xiao and Ashwin Balakrishna and Suraj Nair and Rafael Rafailov and Ethan P Foster and Pannag R Sanketi and Quan Vuong and Thomas Kollar and Benjamin Burchfiel and Russ Tedrake and Dorsa Sadigh and Sergey Levine and Percy Liang and Chelsea Finn},
	booktitle={Conference on Robot Learning},
	year={2024},
}

@inproceedings{zitkovich2023rt,
	title={Rt-2: Vision-language-action models transfer web knowledge to robotic control},
	author={Zitkovich, Brianna and Yu, Tianhe and Xu, Sichun and Xu, Peng and Xiao, Ted and Xia, Fei and Wu, Jialin and Wohlhart, Paul and Welker, Stefan and Wahid, Ayzaan and others},
	booktitle={Conference on Robot Learning},
	pages={2165--2183},
	year={2023},
	organization={PMLR}
}

@inproceedings{
	liu2024rdt,
	title={{RDT}-1B: a Diffusion Foundation Model for Bimanual Manipulation},
	author={Songming Liu and Lingxuan Wu and Bangguo Li and Hengkai Tan and Huayu Chen and Zhengyi Wang and Ke Xu and Hang Su and Jun Zhu},
	booktitle={The Thirteenth International Conference on Learning Representations},
	year={2025}
}

@article{chi2023diffusion,
	title={Diffusion policy: Visuomotor policy learning via action diffusion},
	author={Chi, Cheng and Xu, Zhenjia and Feng, Siyuan and Cousineau, Eric and Du, Yilun and Burchfiel, Benjamin and Tedrake, Russ and Song, Shuran},
	journal={The International Journal of Robotics Research},
	pages={02783649241273668},
	year={2023},
	publisher={SAGE Publications Sage UK: London, England}
}

@inproceedings{belkhale2024rt,
	author={Suneel Belkhale and Tianli Ding and Ted Xiao and Pierre Sermanet and Quan Vuong and Jonathan Tompson and Yevgen Chebotar and Debidatta Dwibedi and Dorsa Sadigh},
	title={RT-H: Action Hierarchies using Language},
	year={2024},
	cdate={1704067200000},
	booktitle={Robotics: Science and Systems},
}

@article{team2024octo,
	title={Octo: An open-source generalist robot policy},
	author={Team, Octo Model and Ghosh, Dibya and Walke, Homer and Pertsch, Karl and Black, Kevin and Mees, Oier and Dasari, Sudeep and Hejna, Joey and Kreiman, Tobias and Xu, Charles and others},
	journal={arXiv preprint arXiv:2405.12213},
	year={2024}
}

@inproceedings{black2024pi_0,
	title={$pi\_0 $: A Vision-Language-Action Flow Model for General Robot Control},
	author={Black, Kevin and Brown, Noah and Driess, Danny and Esmail, Adnan and Equi, Michael and Finn, Chelsea and Fusai, Niccolo and Groom, Lachy and Hausman, Karol and Ichter, Brian and others},
	booktitle={Robotics: Science and Systems},
	year={2025}
}

@article{chen2025training,
	title={Training Strategies for Efficient Embodied Reasoning},
	author={Chen, William and Belkhale, Suneel and Mirchandani, Suvir and Mees, Oier and Driess, Danny and Pertsch, Karl and Levine, Sergey},
	journal={arXiv preprint arXiv:2505.08243},
	year={2025}
}

@article{ma2024survey,
	title={A Survey on Vision-Language-Action Models for Embodied AI},
	author={Ma, Yueen and Song, Zixing and Zhuang, Yuzheng and Hao, Jianye and King, Irwin},
	journal={arXiv preprint arXiv:2405.14093},
	year={2024}
}

@article{xu2024survey,
	title={A survey on robotics with foundation models: toward embodied ai},
	author={Xu, Zhiyuan and Wu, Kun and Wen, Junjie and Li, Jinming and Liu, Ning and Che, Zhengping and Tang, Jian},
	journal={arXiv preprint arXiv:2402.02385},
	year={2024}
}

@inproceedings{kim2025fine,
	title={Fine-tuning vision-language-action models: Optimizing speed and success},
	author={Kim, Moo Jin and Finn, Chelsea and Liang, Percy},
	booktitle={Robotics: Science and Systems},
	year={2025}
}

@article{bjorck2025gr00t,
	title={Gr00t n1: An open foundation model for generalist humanoid robots},
	author={Bjorck, Johan and Casta{\~n}eda, Fernando and Cherniadev, Nikita and Da, Xingye and Ding, Runyu and Fan, Linxi and Fang, Yu and Fox, Dieter and Hu, Fengyuan and Huang, Spencer and others},
	journal={arXiv preprint arXiv:2503.14734},
	year={2025}
}

@article{hung2025nora,
	title={Nora: A small open-sourced generalist vision language action model for embodied tasks},
	author={Hung, Chia-Yu and Sun, Qi and Hong, Pengfei and Zadeh, Amir and Li, Chuan and Tan, U and Majumder, Navonil and Poria, Soujanya and others},
	journal={arXiv preprint arXiv:2504.19854},
	year={2025}
}

@inproceedings{o2024open,
	title={Open x-embodiment: Robotic learning datasets and rt-x models: Open x-embodiment collaboration},
	author={O’Neill, Abby and Rehman, Abdul and Maddukuri, Abhiram and Gupta, Abhishek and Padalkar, Abhishek and Lee, Abraham and Pooley, Acorn and Gupta, Agrim and Mandlekar, Ajay and Jain, Ajinkya and others},
	booktitle={IEEE International Conference on Robotics and Automation},
	pages={6892--6903},
	year={2024},
	organization={IEEE}
}

@inproceedings{walke2023bridgedata,
	title={Bridgedata v2: A dataset for robot learning at scale},
	author={Walke, Homer Rich and Black, Kevin and Zhao, Tony Z and Vuong, Quan and Zheng, Chongyi and Hansen-Estruch, Philippe and He, Andre Wang and Myers, Vivek and Kim, Moo Jin and Du, Max and others},
	booktitle={Conference on Robot Learning},
	pages={1723--1736},
	year={2023},
	organization={PMLR}
}

@INPROCEEDINGS{fang2023rh20t,
	author={Fang, Hao-Shu and Fang, Hongjie and Tang, Zhenyu and Liu, Jirong and Wang, Chenxi and Wang, Junbo and Zhu, Haoyi and Lu, Cewu},
	booktitle={IEEE International Conference on Robotics and Automation}, 
	title={RH20T: A Comprehensive Robotic Dataset for Learning Diverse Skills in One-Shot}, 
	year={2024},
}

@inproceedings{khazatsky2024droid,
	title={DROID: A Large-Scale In-The-Wild Robot Manipulation Dataset},
	author={Khazatsky, Alexander and Pertsch, Karl and Nair, Suraj and Balakrishna, Ashwin and Dasari, Sudeep and Karamcheti, Siddharth and Nasiriany, Soroush and Srirama, Mohan Kumar and Chen, Lawrence Yunliang and Ellis, Kirsty and others},
	booktitle={Robotics: Science and Systems},
	year={2024}
}

@inproceedings{wu2024robomind,
	title={Robomind: Benchmark on multi-embodiment intelligence normative data for robot manipulation},
	author={Wu, Kun and Hou, Chengkai and Liu, Jiaming and Che, Zhengping and Ju, Xiaozhu and Yang, Zhuqin and Li, Meng and Zhao, Yinuo and Xu, Zhiyuan and Yang, Guang and others},
	booktitle={Robotics: Science and Systems}, 
	year={2025}
}

@article{bu2025agibot,
	title={Agibot world colosseo: A large-scale manipulation platform for scalable and intelligent embodied systems},
	author={Bu, Qingwen and Cai, Jisong and Chen, Li and Cui, Xiuqi and Ding, Yan and Feng, Siyuan and Gao, Shenyuan and He, Xindong and Hu, Xuan and Huang, Xu and others},
	journal={arXiv preprint arXiv:2503.06669},
	year={2025}
}

@article{cui2025openhelix,
	title={Openhelix: A short survey, empirical analysis, and open-source dual-system vla model for robotic manipulation},
	author={Cui, Can and Ding, Pengxiang and Song, Wenxuan and Bai, Shuanghao and Tong, Xinyang and Ge, Zirui and Suo, Runze and Zhou, Wanqi and Liu, Yang and Jia, Bofang and others},
	journal={arXiv preprint arXiv:2505.03912},
	year={2025}
}

@article{intelligence2504pi0,
	title={$\pi$0. 5: a vision-language-action model with open-world generalization},
	author={Intelligence, Physical and Black, Kevin and Brown, Noah and Darpinian, James and Dhabalia, Karan and Driess, Danny and Esmail, Adnan and Equi, Michael and Finn, Chelsea and Fusai, Niccolo and others},
	journal={URL https://arxiv. org/abs/2504.16054},
	volume={1},
	number={2},
	pages={3},
	year={2025}
}

@inproceedings{pertsch2025fast,
	title={Fast: Efficient action tokenization for vision-language-action models},
	author={Pertsch, Karl and Stachowicz, Kyle and Ichter, Brian and Driess, Danny and Nair, Suraj and Vuong, Quan and Mees, Oier and Finn, Chelsea and Levine, Sergey},
	booktitle={Robotics: Science and Systems},
	year={2025}
}

@article{song2025accelerating,
	title={Accelerating vision-language-action model integrated with action chunking via parallel decoding},
	author={Song, Wenxuan and Chen, Jiayi and Ding, Pengxiang and Zhao, Han and Zhao, Wei and Zhong, Zhide and Ge, Zongyuan and Ma, Jun and Li, Haoang},
	journal={arXiv preprint arXiv:2503.02310},
	year={2025}
}

@article{team2025robobrain,
	title={RoboBrain 2.0 Technical Report},
	author={Team, BAAI RoboBrain and Cao, Mingyu and Tan, Huajie and Ji, Yuheng and Lin, Minglan and Li, Zhiyu and Cao, Zhou and Wang, Pengwei and Zhou, Enshen and Han, Yi and others},
	journal={arXiv preprint arXiv:2507.02029},
	year={2025}
}

@InProceedings{huang2022inner,
	title = 	 {Inner Monologue: Embodied Reasoning through Planning with Language Models},
	author =       {Huang, Wenlong and Xia, Fei and Xiao, Ted and Chan, Harris and Liang, Jacky and Florence, Pete and Zeng, Andy and Tompson, Jonathan and Mordatch, Igor and Chebotar, Yevgen and Sermanet, Pierre and Jackson, Tomas and Brown, Noah and Luu, Linda and Levine, Sergey and Hausman, Karol and ichter, brian},
	booktitle = 	 {Proceedings of The Conference on Robot Learning},
	pages = 	 {1769--1782},
	year = 	 {2023},
	volume = 	 {205},
	series = 	 {Proceedings of Machine Learning Research},
	publisher =    {PMLR},
}

@article{shao2024deepseekmath,
	title={Deepseekmath: Pushing the limits of mathematical reasoning in open language models},
	author={Shao, Zhihong and Wang, Peiyi and Zhu, Qihao and Xu, Runxin and Song, Junxiao and Bi, Xiao and Zhang, Haowei and Zhang, Mingchuan and Li, YK and Wu, Yang and others},
	journal={arXiv preprint arXiv:2402.03300},
	year={2024}
}

@article{bai2025qwen3,
  title={Qwen3-vl technical report},
  author={Bai, Shuai and Cai, Yuxuan and Chen, Ruizhe and Chen, Keqin and Chen, Xionghui and Cheng, Zesen and Deng, Lianghao and Ding, Wei and Gao, Chang and Ge, Chunjiang and others},
  journal={arXiv preprint arXiv:2511.21631},
  year={2025}
}

@article{zhao2025embodied,
	title={Embodied-R: Collaborative Framework for Activating Embodied Spatial Reasoning in Foundation Models via Reinforcement Learning},
	author={Zhao, Baining and Wang, Ziyou and Fang, Jianjie and Gao, Chen and Man, Fanhang and Cui, Jinqiang and Wang, Xin and Chen, Xinlei and Li, Yong and Zhu, Wenwu},
	journal={arXiv preprint arXiv:2504.12680},
	year={2025}
}

@article{lin2025onetwovla,
	title={OneTwoVLA: A Unified Vision-Language-Action Model with Adaptive Reasoning},
	author={Lin, Fanqi and Nai, Ruiqian and Hu, Yingdong and You, Jiacheng and Zhao, Junming and Gao, Yang},
	journal={arXiv preprint arXiv:2505.11917},
	year={2025}
}

@inproceedings{
	zawalski2024robotic,
	title={Robotic Control via Embodied Chain-of-Thought Reasoning},
	author={Micha{\l} Zawalski and William Chen and Karl Pertsch and Oier Mees and Chelsea Finn and Sergey Levine},
	booktitle={Conference on Robot Learning},
	year={2024},
}

@article{liu2025spatialcot,
	title={SpatialCoT: Advancing Spatial Reasoning through Coordinate Alignment and Chain-of-Thought for Embodied Task Planning},
	author={Liu, Yuecheng and Chi, Dafeng and Wu, Shiguang and Zhang, Zhanguang and Hu, Yaochen and Zhang, Lingfeng and Zhang, Yingxue and Wu, Shuang and Cao, Tongtong and Huang, Guowei and others},
	journal={arXiv preprint arXiv:2501.10074},
	year={2025}
}

@article{tan2025reason,
	title={Reason-rft: Reinforcement fine-tuning for visual reasoning},
	author={Tan, Huajie and Ji, Yuheng and Hao, Xiaoshuai and Lin, Minglan and Wang, Pengwei and Wang, Zhongyuan and Zhang, Shanghang},
	journal={arXiv preprint arXiv:2503.20752},
	year={2025}
}

@article{lu2025vla,
	title={Vla-rl: Towards masterful and general robotic manipulation with scalable reinforcement learning},
	author={Lu, Guanxing and Guo, Wenkai and Zhang, Chubin and Zhou, Yuheng and Jiang, Haonan and Gao, Zifeng and Tang, Yansong and Wang, Ziwei},
	journal={arXiv preprint arXiv:2505.18719},
	year={2025}
}

@article{kim2025robot,
	title={Robot-R1: Reinforcement Learning for Enhanced Embodied Reasoning in Robotics},
	author={Kim, Dongyoung and Park, Sumin and Jang, Huiwon and Shin, Jinwoo and Kim, Jaehyung and Seo, Younggyo},
	journal={arXiv preprint arXiv:2506.00070},
	year={2025}
}

@article{liu2024bidirectional,
	title={Bidirectional Decoding: Improving Action Chunking via Closed-Loop Resampling},
	author={Liu, Yuejiang and Hamid, Jubayer Ibn and Xie, Annie and Lee, Yoonho and Du, Max and Finn, Chelsea},
	year={2025},
	journal={International Conference on Learning Representations}
}

@article{team2023gemini,
  title={Gemini: a family of highly capable multimodal models},
  author={Gemini Team},
    journal={arXiv preprint arXiv:2312.11805},
  year={2023}
}

@article{liu2023libero,
  title={LIBERO: Benchmarking Knowledge Transfer for Lifelong Robot Learning},
  author={Liu, Bo and Zhu, Yifeng and Gao, Chongkai and Feng, Yihao and Liu, Qiang and Zhu, Yuke and Stone, Peter},
  journal={arXiv preprint arXiv:2306.03310},
  year={2023}
}

@article{zheng2024tracevla,
  title={Tracevla: Visual trace prompting enhances spatial-temporal awareness for generalist robotic policies},
  author={Zheng, Ruijie and Liang, Yongyuan and Huang, Shuaiyi and Gao, Jianfeng and Daum{\'e} III, Hal and Kolobov, Andrey and Huang, Furong and Yang, Jianwei},
  journal={arXiv preprint arXiv:2412.10345},
  year={2024}
}

@article{qu2025spatialvla,
  title={Spatialvla: Exploring spatial representations for visual-language-action model},
  author={Qu, Delin and Song, Haoming and Chen, Qizhi and Yao, Yuanqi and Ye, Xinyi and Ding, Yan and Wang, Zhigang and Gu, JiaYuan and Zhao, Bin and Wang, Dong and others},
  journal={arXiv preprint arXiv:2501.15830},
  year={2025}
}

@inproceedings{zhao2025cot,
  title={Cot-vla: Visual chain-of-thought reasoning for vision-language-action models},
  author={Zhao, Qingqing and Lu, Yao and Kim, Moo Jin and Fu, Zipeng and Zhang, Zhuoyang and Wu, Yecheng and Li, Zhaoshuo and Ma, Qianli and Han, Song and Finn, Chelsea and others},
  booktitle={Proceedings of the Computer Vision and Pattern Recognition Conference},
  pages={1702--1713},
  year={2025}
}

@article{zhang2024grape,
  title={Grape: Generalizing robot policy via preference alignment},
  author={Zhang, Zijian and Zheng, Kaiyuan and Chen, Zhaorun and Jang, Joel and Li, Yi and Han, Siwei and Wang, Chaoqi and Ding, Mingyu and Fox, Dieter and Yao, Huaxiu},
  journal={arXiv preprint arXiv:2411.19309},
  year={2024}
}

@article{bu2025univla,
  title={Univla: Learning to act anywhere with task-centric latent actions},
  author={Bu, Qingwen and Yang, Yanting and Cai, Jisong and Gao, Shenyuan and Ren, Guanghui and Yao, Maoqing and Luo, Ping and Li, Hongyang},
  journal={arXiv preprint arXiv:2505.06111},
  year={2025}
}

@article{li2025simplevla,
  title={SimpleVLA-RL: Scaling VLA Training via Reinforcement Learning},
  author={Li, Haozhan and Zuo, Yuxin and Yu, Jiale and Zhang, Yuhao and Yang, Zhaohui and Zhang, Kaiyan and Zhu, Xuekai and Zhang, Yuchen and Chen, Tianxing and Cui, Ganqu and others},
  journal={arXiv preprint arXiv:2509.09674},
  year={2025}
}

@misc{cadene2024lerobot,
    author = {Cadene, Remi and Alibert, Simon and Soare, Alexander and Gallouedec, Quentin and Zouitine, Adil and Palma, Steven and Kooijmans, Pepijn and Aractingi, Michel and Shukor, Mustafa and Aubakirova, Dana and Russi, Martino and Capuano, Francesco and Pascal, Caroline and Choghari, Jade and Moss, Jess and Wolf, Thomas},
    title = {LeRobot: State-of-the-art Machine Learning for Real-World Robotics in Pytorch},
    year = {2024}
}

@article{sapkota2025vision,
  title={Vision-language-action models: Concepts, progress, applications and challenges},
  author={Sapkota, Ranjan and Cao, Yang and Roumeliotis, Konstantinos I and Karkee, Manoj},
  journal={arXiv preprint arXiv:2505.04769},
  year={2025}
}

@ARTICLE{NonAutoregressive,
  author={Xiao, Yisheng and Wu, Lijun and Guo, Junliang and Li, Juntao and Zhang, Min and Qin, Tao and Liu, Tie-Yan},
  journal={IEEE Transactions on Pattern Analysis and Machine Intelligence}, 
  title={A Survey on Non-Autoregressive Generation for Neural Machine Translation and Beyond}, 
  year={2023},
}

@article{chen2025robotwin,
  title={Robotwin 2.0: A scalable data generator and benchmark with strong domain randomization for robust bimanual robotic manipulation},
  author={Chen, Tianxing and Chen, Zanxin and Chen, Baijun and Cai, Zijian and Liu, Yibin and Li, Zixuan and Liang, Qiwei and Lin, Xianliang and Ge, Yiheng and Gu, Zhenyu and others},
  journal={arXiv preprint arXiv:2506.18088},
  year={2025}
}

@article{fei2025libero,
  title={Libero-plus: In-depth robustness analysis of vision-language-action models},
  author={Fei, Senyu and Wang, Siyin and Shi, Junhao and Dai, Zihao and Cai, Jikun and Qian, Pengfang and Ji, Li and He, Xinzhe and Zhang, Shiduo and Fei, Zhaoye and others},
  journal={arXiv preprint arXiv:2510.13626},
  year={2025}
}

@article{cen2025worldvla,
  title={WorldVLA: Towards Autoregressive Action World Model},
  author={Cen, Jun and Yu, Chaohui and Yuan, Hangjie and Jiang, Yuming and Huang, Siteng and Guo, Jiayan and Li, Xin and Song, Yibing and Luo, Hao and Wang, Fan and others},
  journal={arXiv preprint arXiv:2506.21539},
  year={2025}
}

@article{tan2025interactive,
  title={Interactive Post-Training for Vision-Language-Action Models},
  author={Tan, Shuhan and Dou, Kairan and Zhao, Yue and Kr{\"a}henb{\"u}hl, Philipp},
  journal={arXiv preprint arXiv:2505.17016},
  year={2025}
}

@article{chen2025sft,
  title={Sft or rl? an early investigation into training r1-like reasoning large vision-language models},
  author={Chen, Hardy and Tu, Haoqin and Wang, Fali and Liu, Hui and Tang, Xianfeng and Du, Xinya and Zhou, Yuyin and Xie, Cihang},
  journal={arXiv preprint arXiv:2504.11468},
  year={2025}
}

@article{lin2026mmfinereason,
  title={MMFineReason: Closing the Multimodal Reasoning Gap via Open Data-Centric Methods},
  author={Lin, Honglin and Liu, Zheng and Zhu, Yun and Qin, Chonghan and Lin, Juekai and Shang, Xiaoran and He, Conghui and Zhang, Wentao and Wu, Lijun},
  journal={arXiv preprint arXiv:2601.21821},
  year={2026}
}

@inproceedings{song2026maniplvm,
  title={Maniplvm-r1: Reinforcement learning for reasoning in embodied manipulation with large vision-language models},
  author={Song, Zirui and Ouyang, Guangxian and Li, Mingzhe and Ji, Yuheng and Wang, Chenxi and Xu, Zixiang and Zhang, Zeyu and Zhang, Xiaoqing and Jiang, Qian and Ji, Fengxian and others},
  booktitle={Proceedings of the AAAI Conference on Artificial Intelligence},
  volume={40},
  number={22},
  pages={18558--18566},
  year={2026}
}

@article{tan2026robobrain,
  title={RoboBrain 2.5: Depth in Sight, Time in Mind},
  author={Tan, Huajie and Zhou, Enshen and Li, Zhiyu and Xu, Yijie and Ji, Yuheng and Chen, Xiansheng and Chi, Cheng and Wang, Pengwei and Jia, Huizhu and Ao, Yulong and others},
  journal={arXiv preprint arXiv:2601.14352},
  year={2026}
}
\bibliographystyle{colm2026_conference}

%%%%%%%%%%%%%%%%%%%%%%%%%%%%%%%%%%%%%%%%%%%%%%%%%%%%%%%%%%%%%%%%%%%%%%%%%%%%%%%
%%%%%%%%%%%%%%%%%%%%%%%%%%%%%%%%%%%%%%%%%%%%%%%%%%%%%%%%%%%%%%%%%%%%%%%%%%%%%%%
% APPENDIX
%%%%%%%%%%%%%%%%%%%%%%%%%%%%%%%%%%%%%%%%%%%%%%%%%%%%%%%%%%%%%%%%%%%%%%%%%%%%%%%
%%%%%%%%%%%%%%%%%%%%%%%%%%%%%%%%%%%%%%%%%%%%%%%%%%%%%%%%%%%%%%%%%%%%%%%%%%%%%%%
\newpage
\appendix
\section{Appendix}
\subsection{CoT Data Construction Prompt}
\label{appendix:cot_prompt}
	\begin{figure}[htbp]
		\centering
    \begin{tcolorbox}[colback=white, colframe=black!15, arc=2mm, boxrule=0.4pt]
\begin{lstlisting}[style=promptstyle]
Role Assignment
You are an advanced robotic intelligence agent.
Inputs
• Global Task Instruction - overall goal.
• Keyframes - an ordered list of N external-camera images capturing critical moments.
- Frame 1 is before any subtask; Frame i (i > 1) is after subtask i-1 and before subtask i.
Required Output
For each keyframe (in order), output exactly:
Produce exactly N consecutive (<think> ... </think><subtask> ... </subtask>) pairs-one pair per image, same order, plain text, no extra lines. (≤ 50 words, single paragraph, no line breaks)
<think> rules
1. Frame 1: declare initial frame.
2. Frame i (i > 1): internally compare the previous image with the current image to infer the effect of the last subtask, but **do not state Success or Failure explicitly**.
3. Tag every instruction-relevant object with its location in the current image, e.g., bowl (right-front).
4. Describe spatial layout, affordances, obstacles, and reasoning that motivates an immediate next subgoal advancing the Global Task.
5. Do not mention other frames, prior success, failure, or progress metrics. Just analyze the current frame.
6. Do not mention frame indices like Frame 1; use initial frame.
<subtask> rules
• State the immediate next subgoal in clear natural language, **without numerals or explicit quantities**.
• Each subtask must correspond to the change that should occur between the current image and the next one (or to completion at the final image).
• If the Global Task is fulfilled in the current image, output exactly <subtask>finish</subtask> and stop; omit remaining pairs.
Global Constraints
• Output pairs equal exactly the number of keyframes provided.
• No extra text outside the mandated tags.
• No bullet points or lists.
• No numbers inside any <subtask> tag.
• Reasoning and subtasks must align with the images, temporal flow, and the global instruction.
\end{lstlisting}
    \end{tcolorbox}
    \caption{\textbf{Prompt for keyframe CoT annotation via a cloud LVLM.}
        The prompt elicits per-frame spatial scene descriptions and temporal subtask decompositions.}
    \label{data_construct_prompt}
	\end{figure}

    \subsection{Implementation Details}
    \label{appendix:implementation}

    DeepThinkVLA is initialized from the public \(\pi_{0}\)\texttt{-FAST} weights~\citep{pertsch2025fast}.
    We refactor the baseline policy with our hybrid-attention decoder (Sec.~\ref{sec3-2}), yielding a 2.9B parameter model.

    \paragraph{Dataset Construction.}
    Before SFT, we construct embodied CoT datasets using the two-stage pipeline described in Sec.~\ref{sec3-3}, based on public robot manipulation demonstrations.
    Specifically, starting from the LIBERO dataset, this process yields 273{,}465 CoT-annotated frames, which serve as the primary supervision source for the SFT cold-start stage.
    In addition, to support training and evaluation under high-fidelity dynamics and longer task horizons, we apply the same pipeline to RoboTwin~2.0 and construct a large-scale embodied CoT dataset comprising 2{,}847{,}856 annotated frames.
    Together, these datasets provide grounded reasoning supervision across both standard simulation and high-fidelity digital twin environments.

    \paragraph{Supervised Fine-Tuning (SFT).}
    For the SFT stage, we train with a batch size of 128 and a learning rate of \(2.5\times10^{-5}\) for 150k steps.
    A hybrid attention mask is employed so that CoT tokens are supervised autoregressively while action tokens are supervised bidirectionally within the same forward pass.
    Optimization is performed using a standard token-level cross-entropy loss.

    \paragraph{Reinforcement Learning (RL).}
    For the RL stage, we adopt Group Relative Policy Optimization (GRPO)~\citep{shao2024deepseekmath}.
    The action chunk size is set to 10. Each trajectory receives a sparse task-success reward plus a small format-regularization reward to maintain CoT quality.
    Policy updates use a mini-batch size of 128, a low clip ratio \(\epsilon=0.2\), a high clip ratio \(\epsilon=0.28\), and a KL penalty to the SFT reference model to avoid catastrophic forgetting.

    \paragraph{Infrastructure and Inference.}
    Training is conducted on 8$\times$NVIDIA A800 GPUs.
    At inference time, we use greedy decoding for both reasoning and action tokens.
    For CoT ablation experiments, we evaluate three inference modes: Full CoT, Mask CoT, and Random CoT.

\begin{figure}[t!]
        \centering
        \includegraphics[width=0.95\linewidth]{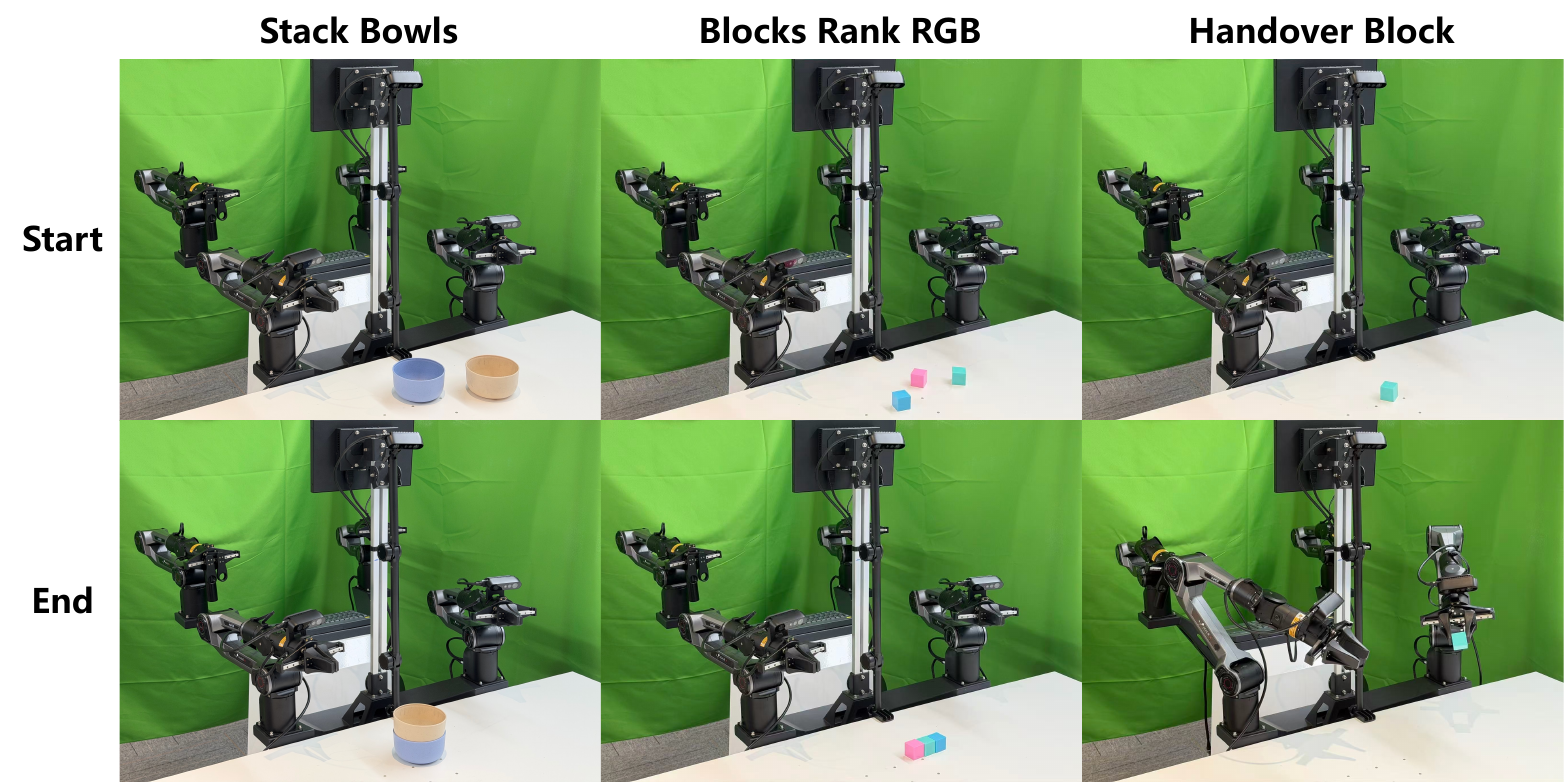}
        \caption{\textbf{Real-robot setup.}
        Start (top row) and end (bottom row) configurations of the three physical manipulation tasks---Stack Bowls, Blocks Rank RGB, and Handover Block---on the AGILEX ALOHA bimanual platform with external cameras.}
        \label{fig:real_setup}
    \end{figure}

    \paragraph{Real-Robot Setup and Calibrated Comparison.}
    \label{appendix:real_robot_details}
    For the real-robot experiments (Section~\ref{sec:real_robot}), we use an AGILEX ALOHA bimanual robot platform equipped with external cameras; Figure~\ref{fig:real_setup} shows the physical setup together with the start and end configurations of each task.
    A human teleoperator collects demonstration trajectories for each of the three tasks, which span a range of manipulation skills and difficulty levels: Stack Bowls Two requires stacking two bowls into a nested configuration with precise alignment; Handover Block transfers a block between grippers, testing bimanual coordination; Blocks Rank RGB arranges colored blocks (red, green, blue) in a specified order, requiring color recognition and sequential reasoning.

    We apply the identical two-stage CoT annotation pipeline (Section~\ref{sec3-3}) to the collected demonstrations: keyframes identified by gripper state changes are annotated using a cloud-based VLM, and a locally fine-tuned model propagates CoT to intermediate frames.
    The model is initialized from the same $\pi_0$-FAST weights and fine-tuned via SFT on the resulting real-robot CoT dataset.
    Evaluation is conducted over 20 trials per task with randomized initial object configurations.

    The physical study deliberately uses SFT only. Performing online RL on the physical robot would require many trial-and-error rollouts---including failed grasps and contact-rich mistakes---which is unsafe without dedicated online-RL safety infrastructure; claims about RL-based causal alignment are therefore grounded in the controlled simulation and RoboTwin~2.0 experiments, where on-policy rollouts can be collected safely and at scale.
    Moreover, the 45\% average success should not be compared directly with the near-saturated LIBERO results; a more appropriate reference is the harder, contact-rich RoboTwin~2.0 benchmark (Table~\ref{tab:robotwin}), where the simulation results on the three corresponding task categories are 62.0\% (Stack Bowls Two), 43.0\% (Handover Block), and 77.0\% (Blocks Rank RGB).
    The physical results are lower---most notably on Blocks Rank RGB---but remain within the same order of difficulty rather than collapsing from LIBERO-level performance, showing that the CoT annotation pipeline and hybrid decoding architecture can be instantiated on hardware and achieve nontrivial closed-loop execution under real camera, calibration, contact, and actuation noise.
    We emphasize the scope of this evidence: it supports the preliminary physical applicability of the CoT data construction and hybrid action decoder; it does not validate the full SFT-then-RL pipeline in the physical domain.

    \subsection{Hybrid Decoder: Token-Level Specification}
    \label{appendix:hybrid_decoder}

    This section specifies the hybrid decoder of Section~\ref{sec3-2} at the token-flow level, covering how action chunk slots are represented, how ground-truth leakage is prevented, and how the attention mask is implemented.

    \paragraph{Sequence Layout and Action Tokenization.}
    Each policy call follows the sequence layout: observation/instruction prompt, \texttt{<think>} CoT tokens \texttt{</think>}, \texttt{<action>}, an action chunk, \texttt{</action>}, and EOS. The \texttt{<think>...</think>} delimiters mark the natural-language reasoning segment. The \texttt{<action>} and \texttt{</action>} tokens are only delimiters for the action segment: the content inside is not natural language but a fixed block of scalar action-value positions. Each scalar action value is normalized to $[-1,1]$ and discretized by the \texttt{ActionTokenizer} into one of 2{,}048 action-token IDs reused from the tail of the base tokenizer vocabulary; these are reused vocabulary IDs, not newly added per-bin string tokens. With \texttt{n\_bins}=2048 and \texttt{fast\_skip\_tokens}=128, the action-token ID range is $[\texttt{vocab\_size}-128-2048,\ \texttt{vocab\_size}-128-1]$; in our released base tokenizer with \texttt{vocab\_size}=257152, this is $[254976, 257023]$. The final 128 base-vocabulary IDs are skipped, and \texttt{<action>}/\texttt{</action>} are separate delimiter special tokens. The mapping is reversed: discrete bin index $k$ is encoded as token ID $\texttt{vocab\_size}-1-128-k$. The action block has $H \times d_a$ positions: the position index specifies the action timestep and action dimension, while the predicted token ID specifies the discretized scalar value for that coordinate. For LIBERO, $H=10$ and $d_a=7$, so each policy call contains 70 scalar action positions; RoboTwin uses the same construction with 14-D actions.

    \paragraph{Action Slots: Neither Mask Tokens Nor Learned Queries.}
    The action positions are not learned query tokens and are not semantic mask tokens. During SFT, the labels contain the ground-truth action-value tokens, but before the transformer forward pass, the input embeddings at action-token positions are zeroed by the action-position mask. During inference, we append dummy placeholder IDs only to allocate the same $H \times d_a$ positions, and these placeholder embeddings are likewise zeroed. The slots therefore preserve positional structure but do not reveal ground-truth action embeddings.

    \paragraph{Hybrid Attention Mask and Parallel Decoding.}
    Prompt and visual tokens are visible as context; the \texttt{<think>...</think>} reasoning segment is generated with the standard causal language-model mask. From the \texttt{<action>} region onward, the action-slot rows copy a fully visible valid-sequence row, so every action slot can attend to the prompt, the generated CoT, and all other action slots. At inference time, the model first autoregressively generates the reasoning segment up to \texttt{</think>} and \texttt{<action>}; we then append the zero-input action slots, \texttt{</action>}, and EOS, run one transformer forward pass with the hybrid mask, restrict the LM head to the reserved action-token range, and decode all $H \times d_a$ action values in parallel. Figure~\ref{fig:hybrid_decoder_pseudo} summarizes placeholder construction, embedding zeroing, and action-row mask replacement.

    \begin{figure}[t!]
        \centering
        \begin{tcolorbox}[colback=white, colframe=black!15, arc=2mm, boxrule=0.4pt]
\begin{lstlisting}[style=pseudostyle]
# Notation: H = action-chunk length, d_a = action dimension,
#   V = vocab_size (257152), S = fast_skip_tokens (128), B = n_bins (2048)
# Reserved action-token slice (reused tail of the base vocabulary):
#   begin = V - S - B                       # 254976
#   end   = V - S - 1                       # 257023
#   encode(bin k) -> id = V - 1 - S - k ;  decode(id) -> bin = V - 1 - S - id

# 1) Placeholder construction: allocate H x d_a action positions
seq = [prompt, <think>, cot, </think>, <action>, act_slots(H x d_a), </action>, EOS]
#   SFT:       act_slots hold ground-truth action-value ids (labels only)
#   inference: act_slots hold dummy placeholder ids

# 2) Embedding zeroing: slots keep positional structure, leak no action values
emb = embed(seq)
emb[action_positions] = 0

# 3) Hybrid attention mask: action-row replacement
mask = causal_mask(len(seq))                # prompt + CoT remain causal
for row in action_positions:
    mask[row, :valid_len] = 1               # action rows: fully visible

# 4) Training: one forward pass, token-level cross-entropy
logits = transformer(emb, mask)             # CoT causal, action slots bidirectional

# 5) Inference
cot = autoregressive_decode(prompt, until=</think><action>)   # causal attention
seq = append(cot, zeroed_act_slots, </action>, EOS)
logits = transformer(embed_and_zero(seq), hybrid_mask)        # one forward pass
action_logits = logits[action_positions, begin:end+1]         # restrict LM head
actions = detokenize(argmax(action_logits))                   # H x d_a values in parallel
\end{lstlisting}
        \end{tcolorbox}
        \caption{\textbf{Pseudocode for the hybrid decoder.}
        Placeholder construction allocates $H \times d_a$ zero-input action slots; embedding zeroing prevents ground-truth leakage during SFT; action-row mask replacement grants action slots bidirectional visibility over the prompt, generated CoT, and all other slots, enabling one-pass parallel action decoding.}
        \label{fig:hybrid_decoder_pseudo}
    \end{figure}

    \subsection{Baselines Details}
    \label{appendix:baselines}

    We compare DeepThinkVLA with a comprehensive set of Vision--Language--Action (VLA) baselines used across our three benchmarks: LIBERO~\citep{liu2023libero}, LIBERO-Plus~\citep{fei2025libero}, and RoboTwin~2.0~\citep{chen2025robotwin}. The baselines span autoregressive policies, diffusion/flow-based policies, and parallel/block-decoding variants. Unless otherwise specified, results are reported following the official evaluation protocols of each benchmark.

    \paragraph{Autoregressive VLA baselines (LIBERO).}
    We include representative autoregressive (AR) VLAs that map vision and language to action tokens sequentially.
    TraceVLA~\citep{zheng2024tracevla} augments AR decoding with spatial-temporal trace cues from past trajectories.
    Octo~\citep{team2024octo} is a generalist AR policy trained across diverse manipulation data.
    OpenVLA~\citep{kim2024openvla} is a widely adopted open-source AR baseline for language-conditioned manipulation.
    SpatialVLA~\citep{qu2025spatialvla} incorporates explicit spatial modeling to improve AR policies.
    NORA~\citep{hung2025nora} is an efficient AR VLA with strong performance on standard manipulation suites.
    UniVLA~\citep{bu2025univla} trains a unified AR policy across multiple task suites.
    $\pi_{0}$\texttt{-FAST}~\citep{pertsch2025fast} is a strong AR baseline based on the FAST action tokenizer.
    We also include AR policies further optimized beyond static imitation:
    GRAPE~\citep{zhang2024grape} introduces RL-style optimization on top of a base policy.
    VLA-RL~\citep{lu2025vla} applies outcome-driven RL to improve task success over an AR backbone.

    \paragraph{Diffusion / flow-based policies (LIBERO and RoboTwin 2.0).}
    Diffusion-based policies generate continuous actions through iterative refinement.
    Diffusion Policy~\citep{chi2023diffusion} is a standard diffusion baseline widely used in robot learning.
    $\pi_{0}$~\citep{black2024pi_0} is a strong flow/diffusion-style VLA that serves as a competitive non-autoregressive baseline and is evaluated in our LIBERO, LIBERO-Plus, and RoboTwin~2.0 comparisons.

    \paragraph{Parallel / block-decoding policies (LIBERO).}
    Parallel or block-decoding methods use bidirectional attention to predict multiple action tokens jointly, reducing latency and mitigating autoregressive interference.
    CoT-VLA-7B~\citep{zhao2025cot} supports Chain-of-Thought supervision with block-parallel action decoding.
    OpenVLA--OFT~\citep{kim2025fine} is a bidirectional/parallel-decoding variant built on OpenVLA, designed to improve inference efficiency and stability.

    \paragraph{Baselines specific to LIBERO-Plus (robustness benchmark).}
    LIBERO-Plus~\citep{fei2025libero} evaluates robustness under seven perturbation families (camera, robot initialization, language rewrites, lighting, background, sensor noise, and object layout).
    We report the official robustness results for the following baselines included in Table~\ref{tab:perturb_robustness}:
    $\pi_{0}$~\citep{black2024pi_0},
    $\pi_{0}$\texttt{-FAST}~\citep{pertsch2025fast},
    UniVLA~\citep{bu2025univla},
    WorldVLA~\citep{cen2025worldvla},
    OpenVLA~\citep{kim2024openvla},
    OpenVLA--OFT~\citep{kim2025fine},
    NORA-long~\citep{hung2025nora},
    and RIPT-VLA~\citep{tan2025interactive}.
    For clarity, we follow the benchmark conventions and additionally include two commonly reported OpenVLA--OFT variants:
    OpenVLA--OFT\_w (augmented with wrist-camera observations) and OpenVLA--OFT\_m (trained with additional mixed data), as shown in Table~\ref{tab:perturb_robustness}.

    \paragraph{Baselines specific to RoboTwin 2.0 (high-fidelity digital twin).}
    RoboTwin~2.0~\citep{chen2025robotwin} features longer horizons and contact-rich dynamics.
    We compare against widely used baselines reported in Table~\ref{tab:robotwin}:
    $\pi_{0}$~\citep{black2024pi_0},
    $\pi_{0}$\texttt{-FAST}~\citep{pertsch2025fast},
    RDT~\citep{liu2024rdt},
    and OpenVLA--OFT~\citep{kim2025fine}.
    These cover both diffusion/flow-based policies and parallel-decoding approaches under the same benchmark protocol.

    \begin{table}[t!]
        \centering
        \caption{\textbf{Ablation on CoT and hybrid-decoding} (success rate \% and relative inference time).
        Naively adding CoT to an AR decoder hurts performance; hybrid decoding restores and amplifies the benefit.}
        \label{CoT hybrid compare}
        \renewcommand{\arraystretch}{1.2}
        \resizebox{\linewidth}{!}{
            \begin{tabular}{llcccccl}
                \toprule
                \textbf{Category} & \textbf{Method} & \textbf{Object (\%)} & \textbf{Spatial (\%)} & \textbf{Goal (\%)} & \textbf{Long (\%)} & \textbf{Average (\%)} & \textbf{Rel.\ Inference Time} \\
                \midrule
                \multirow{1}{*}{Baseline}
                    & $\pi_{0}$\texttt{-FAST} & 96.8 & 96.4 & 88.6 & 60.2 & 85.5 & \quad\quad$1.0\times$ \\
                \midrule
                \multirow{1}{*}{AR-CoT}
                    & $\pi_{0}$\texttt{-FAST} (Full CoT) & 95.8 & 93.8 & 74.6 & 61.0 & 81.3 & \quad\quad$4.0\times$ \\
                \midrule
                \multirow{3}{*}{Hybrid CoT}
                    & DeepThinkVLA (Mask CoT) & 99.0 & 97.2 & 96.0 & 93.6 & 96.5 & \quad\quad$0.175\times$ \\
                    & DeepThinkVLA (Random CoT) & 97.8 & 94.4 & 60.2 & 87.8 & 85.1 & \quad\quad$0.175\times$ \\
                    & DeepThinkVLA (Full CoT) & \textbf{99.0} & \textbf{97.2} & \textbf{96.8} & \textbf{94.2} & \textbf{96.8} & \quad\quad$1.4\times$ \\
                \bottomrule
            \end{tabular}
        }
    \end{table}

    \subsection{Detailed Robustness Analysis}
    \label{appendix:robustness}

    \paragraph{Wrist Camera Ablation.}
    Incorporating a wrist camera provides additional improvements by capturing near-field contact information. However, even without the wrist camera, DeepThinkVLA achieves 86.0\% on LIBERO, already outperforming $\pi_0$-FAST (85.5\%). This confirms that the gains primarily stem from the hybrid architecture and reasoning capability rather than specific sensor modalities.

    \paragraph{Deconstructing LIBERO-Plus Robustness.}
    The LIBERO-Plus gains (Table~\ref{tab:perturb_robustness}) are broad-based rather than driven by a single factor. DeepThinkVLA improves camera robustness from 65.1\% to 88.5\%, robot-initial robustness from 21.6\% to 40.5\%, language robustness from 61.0\% to 84.5\%, and noise robustness from 74.4\% to 94.4\%. These results indicate that the hybrid decoding and embodied CoT reasoning help maintain functional competence under substantial perception/instruction-side distribution shifts.

    \paragraph{Perception-Side vs.\ Action-Side OOD: Effect of RL Alignment.}
    The seven LIBERO-Plus dimensions primarily perturb the \emph{perception/instruction side} (camera view, robot initialization, language rewrites, lighting, background, sensor noise, and object layout): they mainly stress the visual-language encoder's invariance and grounding robustness, since a correctly grounded subgoal remains essentially unchanged under, e.g., a camera or lighting change. The Joint-Limit intervention instead perturbs the \emph{action/dynamics side}: the same subgoal plan must be realized by different action trajectories and recovery behavior, so the quality of the CoT-to-action link directly matters. This distinction is why Joint-Limit serves as the designed causal test for Condition~2, and it delimits the scope of our claim: Causal Alignment is required for CoT to remain functional in action-execution-sensitive OOD settings, not for every OOD factor.

    To verify that RL alignment does not sacrifice perception-side robustness, we evaluate the SFT-only and RL-aligned models on every LIBERO-Plus perturbation dimension with no adaptation (Table~\ref{tab:libero_plus_rl}); the DeepThinkVLA ($\pi_0$-FAST) row of Table~\ref{tab:perturb_robustness} corresponds to the SFT-only model. RL alignment preserves the broad zero-shot robustness (79.1\% vs.\ 79.0\% overall), with per-dimension changes confined to a narrow band. We interpret this as a positive robustness check rather than causal evidence: aligning the CoT-to-action pathway neither requires nor damages the perception-side robustness measured by LIBERO-Plus.

    \begin{table}[t!]
        \centering
        \caption{\textbf{LIBERO-Plus perception/instruction-side robustness: SFT-only vs.\ RL-aligned} (success rate \%, zero-shot, no adaptation). RL alignment preserves broad robustness under perception/instruction-side shifts; the causal evidence for Condition~2 remains the action/dynamics-side Joint-Limit study in Table~\ref{tab:libero_ood_mask}.}
        \label{tab:libero_plus_rl}
        \scriptsize
        \setlength{\tabcolsep}{3pt}
        \renewcommand{\arraystretch}{1.05}
        \resizebox{0.99\linewidth}{!}{
        \begin{tabular}{lcccccccc}
        \toprule
        \textbf{Model} & \textbf{Camera} & \textbf{Robot-Init} & \textbf{Language} & \textbf{Light} & \textbf{Background} & \textbf{Noise} & \textbf{Layout} & \textbf{Total} \\
        \midrule
        DeepThinkVLA ($\pi_0$-FAST, SFT-Only)   & 88.5 & 40.5 & 84.5 & 90.0 & 75.3 & 94.4 & 79.9 & 79.0 \\
        DeepThinkVLA ($\pi_0$-FAST, RL-Aligned) & 88.7 & 41.3 & 83.9 & 90.4 & 76.2 & 93.9 & 79.4 & 79.1 \\
        \midrule
        $\Delta$ (RL $-$ SFT) & +0.2 & +0.8 & $-$0.6 & +0.4 & +0.9 & $-$0.5 & $-$0.5 & +0.1 \\
        \bottomrule
        \end{tabular}
        }
    \end{table}

    The OOD Joint-Limit experiment (Table~\ref{tab:libero_ood_mask}) provides the strongest evidence for Condition~2. Despite high standard performance, the SFT-only model mirrors the fragility of the reasoning-free baseline (32.0\,pp vs.\ 31.6\,pp drop)---SFT-learned CoT is ``fake reasoning'' that mimics annotation style without causal influence on actions. Only after RL does the drop shrink to 24.4\,pp, and masking CoT widens it to 27.7\,pp, indicating that RL alignment makes the CoT pathway action-relevant and turns CoT into a functional planning signal under dynamics shift.

    \begin{table*}[tb]
        \centering
        \caption{\textbf{Controlled OOD analysis via Joint-Limit dynamics.}
        SFT-only drops as much as the reasoning-free baseline ($\sim$32\,pp); RL alignment reduces the drop to 24.4\,pp, and masking CoT widens it again, indicating that RL-aligned reasoning actively aids OOD adaptation.
        }
        \label{tab:libero_ood_mask}
        \setlength{\tabcolsep}{11.5pt}
        \resizebox{0.95\linewidth}{!}{
        \begin{tabular}{c|c|c|cccc|c|c}
        \toprule
        \multirow{2}{*}{\textbf{Method / Stage}} & \multirow{2}{*}{\textbf{Inference Mode}} & \multirow{2}{*}{\textbf{Dynamics}} & \multicolumn{5}{c|}{\textbf{Success Rate (\%)}} & \textbf{Drop} \\
        \cmidrule(lr){4-8}
         & & & \textbf{Obj} & \textbf{Spa} & \textbf{Goal} & \textbf{Long} & \textbf{Avg} & \textbf{(pp)} \\
        \midrule
        \multirow{2}{*}{$\pi_0$-FAST \cite{pertsch2025fast}} & \multirow{2}{*}{-} & Standard & 96.8 & 96.4 & 88.6 & 60.2 & 85.5 & - \\
         & & OOD Limit & 77.0 & 64.0 & 31.2 & 43.6 & 53.9 & \textcolor{red}{31.6} \\
        \midrule
        \multirow{4}{*}{DeepThinkVLA(SFT-Only)} & \multirow{2}{*}{Full CoT} & Standard & 99.0 & 97.2 & 96.8 & 94.2 & \textbf{96.8} & - \\
         & & OOD Limit & 87.0 & 61.4 & 54.2 & 56.4 & 64.8 & \textcolor{red}{32.0} \\
        \cmidrule{2-9}
         & \multirow{2}{*}{Mask CoT} & Standard & 99.0 & 97.2 & 95.8 & 93.4 & 96.4 & - \\
         & & OOD Limit & 85.2 & 58.0 & 50.6 & 48.2 & 60.5 & \textcolor{red}{35.9} \\
        \midrule
        \multirow{4}{*}{\textbf{DeepThinkVLA(RL-Aligned)}} & \multirow{2}{*}{\textbf{Full CoT}} & Standard & \textbf{99.0} & \textbf{96.6} & \textbf{96.4} & \textbf{96.2} & \textbf{97.0} & - \\
         & & \textbf{OOD Limit} & \textbf{91.6} & \textbf{64.2} & \textbf{66.2} & \textbf{68.4} & \textbf{72.6} & \textbf{\textcolor{blue}{24.4}} \\
        \cmidrule{2-9}
         & \multirow{2}{*}{Mask CoT} & Standard & 99.0 & 97.2 & 96.0 & 93.6 & 96.5 & - \\
         & & OOD Limit & 88.2 & 60.2 & 61.0 & 65.6 & 68.8 & \textcolor{gray}{27.7} \\
        \bottomrule
        \end{tabular}
        }
    \end{table*}

    \begin{figure}[htbp]
      \centering
      \includegraphics[width=\columnwidth]{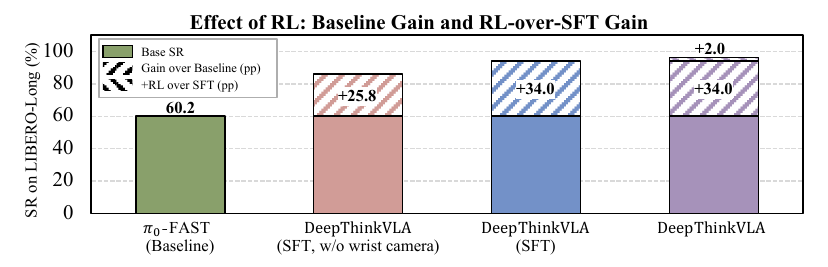}
      \caption{\textbf{Effect of RL on LIBERO-Long.}
      Solid bars share the $\pi_{0}$\texttt{-FAST} baseline (60.2\%); shaded segments show gains from architecture and RL. The teal segment marks the +2\,pp RL improvement over SFT-only.
      }
      \label{fig:rl_libero_long}
    \end{figure}

    \subsection{Ablations on Architecture and the Role of CoT}
    \label{sec:appendix_arch}
    
    \paragraph{Validating Condition~1: Decoding Alignment.}
    To directly test whether Decoding Alignment is necessary for CoT to be effective, we compare a naive autoregressive CoT baseline with our hybrid-decoding architecture.
    Specifically, we apply identical embodied CoT supervision to the autoregressive $\pi_{0}$\texttt{-FAST} model and evaluate against DeepThinkVLA under the same training conditions.
    
    As shown in Table~\ref{CoT hybrid compare}, directly introducing CoT into a fully autoregressive decoder degrades performance (Average: 81.3\% vs.\ 85.5\% without CoT) and incurs a $4\times$ increase in inference latency.
    This is the direct evidence for Condition~1: forcing a single autoregressive stream to jointly model sequential reasoning and high-dimensional actions creates destructive interference between modalities, making CoT actively harmful.
    In contrast, the hybrid-decoding architecture consistently achieves substantially higher success rates (Average: 96.8\%) with only modest latency overhead, yielding a +15.5 percentage point improvement over the naive AR-CoT baseline.
    These results confirm that Decoding Alignment---matching the generation mechanism to each modality's intrinsic structure---is a necessary condition for CoT to benefit VLA models.

    \paragraph{Semantic Coherence of CoT.}
    To examine whether CoT functions as an explicit reasoning signal rather than a mere auxiliary token stream, we further analyze its semantic role at inference time.
    We consider two controlled variants: Mask CoT, where reasoning tokens are replaced by placeholders, and Random CoT, where CoT tokens are substituted with randomly generated content.

    As reported in Table~\ref{CoT hybrid compare}, Mask CoT yields only a minor performance change under standard evaluation (Average: 96.5\%), reflecting near-saturation in-distribution.
    In contrast, Random CoT results in a substantial degradation (Average: 85.1\%), demonstrating that once reasoning tokens are consumed by the policy, their semantic coherence is critical for successful execution.
    The apparently conflicting pattern---Mask CoT $\approx$ Full CoT yet Random CoT $\ll$ Full CoT---is best explained by a \emph{dual-channel} behavior: the policy conditions on a coherent CoT when it is present, but falls back to a strong visual-action prior when the CoT signal is absent.
    Consistent with this view, under out-of-distribution dynamics (Table~\ref{tab:libero_ood_mask}) the visual-action prior becomes less sufficient---the same subgoal must be realized by a different execution trajectory---and the gap between Full CoT (72.6\%) and Mask CoT (68.8\%) widens accordingly.
    We therefore state the role of CoT precisely: aligned CoT is not uniformly important on every in-distribution step; it functions as a planning signal that is most measurably useful when the visual-action prior is insufficient, such as under dynamics shift, error recovery, and long-horizon execution.

    \paragraph{Same-Architecture Control: Removing CoT from Training.}
    The Mask/Random CoT interventions above operate at inference time with fixed weights. As a complementary training-time control, we retrain DeepThinkVLA \emph{without} the CoT pathway while keeping the action decoder and the RL alignment procedure fixed: during SFT, CoT supervision is removed; during RL rollouts and inference, CoT generation is skipped. This control separates the contribution of the explicit CoT pathway from that of the hybrid action decoder and the RL training stage, and directly tests whether the final performance can be explained by the generic SFT-then-RL recipe alone.

    As shown in Table~\ref{tab:no_cot_control} (in the main text), the w/o-CoT model reaches 92.7\% after SFT and 93.8\% after RL, confirming that the hybrid decoder and the RL pipeline are effective on their own. However, the full model with CoT under the identical RL procedure reaches 97.0\%: a further +3.2\,pp that cannot be attributed to the action decoder or to the generic SFT-then-RL recipe, with especially large gains on Goal (+4.6\,pp) and Long (+6.0\,pp).
    This control also fixes the data pipeline, task distribution, and CoT annotation quality; it therefore isolates what the explicit CoT pathway contributes at this data-quality level, while higher-quality CoT annotation remains an orthogonal axis that may further raise performance~\citep{chen2025sft, lin2026mmfinereason}.
    Together with the fixed-weight Mask/Random CoT interventions and the OOD analysis in Appendix~\ref{appendix:robustness}, these results provide converging functional evidence---from training-time removal, fixed-weight masking, random semantic corruption, and dynamics-shift robustness---that the CoT pathway is functionally useful for action generation. We note this is functional evidence, not a complete mechanistic proof of human-like reasoning.

    \begin{figure}[t!]
        \centering
        \includegraphics[width=0.8\linewidth]{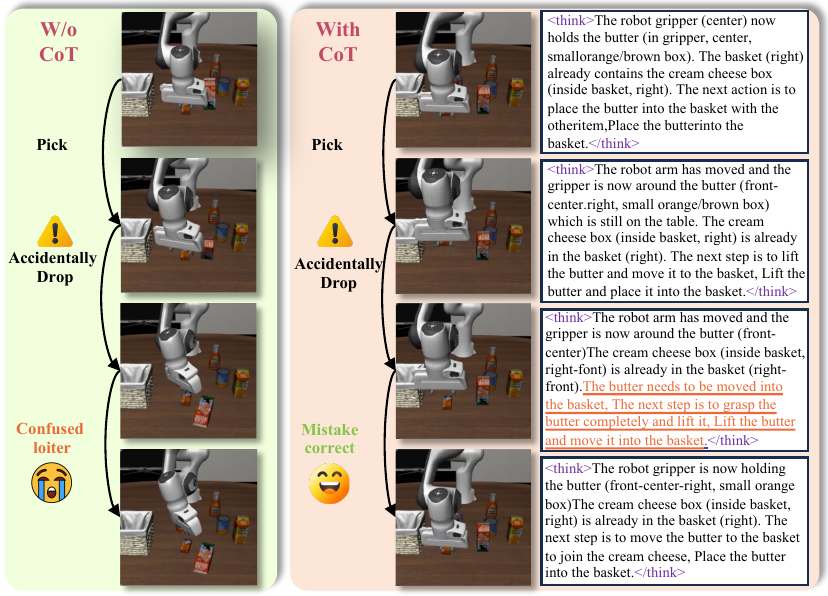}
        \caption{\textbf{``Think before acting'' enables error recovery.}
        \textbf{Left:} the baseline enters a repetitive failure loop after a missed grasp.
        \textbf{Right:} DeepThinkVLA re-articulates the subgoal via CoT and recovers.}
        \label{cot-self-correct}
    \end{figure}

    \subsection{CoT Quality Audits}
    \label{appendix:cot_audit}

    Because dataset scale alone does not guarantee supervision quality, we report two complementary text-level audits: a \emph{demonstration-data audit} that checks the CoT annotations used as SFT supervision, and a \emph{deployment audit} that checks whether the trained policy still generates structured CoTs at test time. Both audits are necessary-not-sufficient checks: they measure text form, not the functional use of CoT, which is established separately in Appendix~\ref{sec:appendix_arch}.

    \paragraph{Audit Protocol.}
    Each sample is first parsed for exactly one nonempty \texttt{<think>...</think>} block; malformed blocks are counted as malformed and receive 0 on the five pass checks. For valid blocks, the criteria are deterministic and intentionally broad: \emph{object grounding} requires at least one task/allowed object mention; \emph{spatial relation} checks spatial terms; \emph{temporal subgoal} checks task/action verbs; \emph{no unexpected object} rejects unexpected object phrases; and \emph{not low-level transcript} requires object mentions, at least 12 alphabetic words, and a low-level action/coordinate-term ratio below 0.08. The deployment audit adds two deliberately stricter planning-marker checks on the same generated CoTs: the first passes when the text contains terms from at least two manipulation-stage groups (approach, grasp, transport, place, operate); the second passes when the text contains an explicit phrase linking the task goal or scene state to the next subgoal.

    \paragraph{Demonstration-Data Audit.}
    We sample 1{,}000 CoT annotations, balanced across 500 LIBERO demonstration CoTs (from a 40-task stratified audit pool) and 500 RoboTwin demonstration CoTs (from 24 random episodes). As shown in Table~\ref{tab:cot_audit_demo}, the CoT supervision is generally object-grounded, spatially specific, and subgoal-oriented rather than an action transcript: 100.0\% pass object grounding, 99.9\% pass temporal subgoal description, 100.0\% pass the not-low-level-transcript check, and no malformed tags occur. We conservatively count any unexpected object mention as a failure, which gives 96.8\% overall on that criterion.

    \begin{table}[t!]
        \centering
        \caption{\textbf{Demonstration-data CoT quality audit} (pass counts and rates over 1{,}000 sampled annotations). The supervision is object-grounded, spatially specific, and subgoal-oriented rather than a low-level action transcript; unexpected object mentions are conservatively counted as failures.}
        \label{tab:cot_audit_demo}
        \scriptsize
        \setlength{\tabcolsep}{4pt}
        \renewcommand{\arraystretch}{1.05}
        \resizebox{0.99\linewidth}{!}{
        \begin{tabular}{lccc}
        \toprule
        \textbf{Quality criterion} & \textbf{LIBERO demos ($N$=500)} & \textbf{RoboTwin demos ($N$=500)} & \textbf{Overall ($N$=1000)} \\
        \midrule
        Object grounding & 500 / 500 (100.0\%) & 500 / 500 (100.0\%) & 1000 / 1000 (100.0\%) \\
        Spatial relation description & 500 / 500 (100.0\%) & 500 / 500 (100.0\%) & 1000 / 1000 (100.0\%) \\
        Temporal subgoal description & 499 / 500 (99.8\%) & 500 / 500 (100.0\%) & 999 / 1000 (99.9\%) \\
        No hallucinated/unexpected object mention & 468 / 500 (93.6\%) & 500 / 500 (100.0\%) & 968 / 1000 (96.8\%) \\
        Not low-level action transcript & 500 / 500 (100.0\%) & 500 / 500 (100.0\%) & 1000 / 1000 (100.0\%) \\
        Malformed tag rate & 0 / 500 (0.0\%) & 0 / 500 (0.0\%) & 0 / 1000 (0.0\%) \\
        \bottomrule
        \end{tabular}
        }
    \end{table}

    \paragraph{Deployment Audit.}
    We further sample 1{,}000 generated CoTs from the official RL-aligned checkpoint during deployment inference: 500 from standard inference and 500 from Joint-Limit inference, covering 10 tasks per regime with 50 generated CoTs per task. As shown in Table~\ref{tab:cot_audit_deploy}, generated CoTs remain object-grounded, spatially specific, and subgoal-oriented in deployment, including under Joint-Limit inference: the broad non-degeneration checks pass at 99.4--99.9\% overall, and malformed tags occur in 0.1\% of samples. The stricter rows are intentionally discriminative and non-saturated: 36.7\% of generated CoTs contain terms from at least two manipulation-stage groups, and 38.2\% contain an explicit goal-to-next-step phrase.

    \begin{table}[t!]
        \centering
        \caption{\textbf{Deployment generated-CoT text audit} over 1{,}000 CoTs generated by the RL-aligned checkpoint. Upper rows: broad format/non-degeneration checks. Lower rows: deliberately stricter planning-marker checks on the same generated CoTs.}
        \label{tab:cot_audit_deploy}
        \scriptsize
        \setlength{\tabcolsep}{4pt}
        \renewcommand{\arraystretch}{1.05}
        \resizebox{0.99\linewidth}{!}{
        \begin{tabular}{lccc}
        \toprule
        \textbf{Format / non-degeneration check} & \textbf{Standard inference ($N$=500)} & \textbf{Joint-Limit inference ($N$=500)} & \textbf{Overall ($N$=1000)} \\
        \midrule
        Contains task/scene object reference & 500 / 500 (100.0\%) & 499 / 500 (99.8\%) & 999 / 1000 (99.9\%) \\
        Contains spatial relation term & 500 / 500 (100.0\%) & 499 / 500 (99.8\%) & 999 / 1000 (99.9\%) \\
        Contains manipulation/subgoal verb & 500 / 500 (100.0\%) & 499 / 500 (99.8\%) & 999 / 1000 (99.9\%) \\
        No unexpected object mention & 498 / 500 (99.6\%) & 496 / 500 (99.2\%) & 994 / 1000 (99.4\%) \\
        Not low-level action transcript & 500 / 500 (100.0\%) & 499 / 500 (99.8\%) & 999 / 1000 (99.9\%) \\
        Malformed tag rate & 0 / 500 (0.0\%) & 1 / 500 (0.2\%) & 1 / 1000 (0.1\%) \\
        \midrule
        \multicolumn{4}{l}{\textit{Stricter planning-marker checks on the same 1{,}000 generated CoTs}} \\
        Contains $\geq$2 manipulation-stage groups & 200 / 500 (40.0\%) & 167 / 500 (33.4\%) & 367 / 1000 (36.7\%) \\
        Contains explicit goal-to-next-step phrase & 188 / 500 (37.6\%) & 194 / 500 (38.8\%) & 382 / 1000 (38.2\%) \\
        \bottomrule
        \end{tabular}
        }
    \end{table}

    We interpret these audits conservatively. The text evidence supports object-grounded next-subgoal planning with a nontrivial subset of explicit temporal/planning markers; it is not, by itself, proof of human-like reasoning or deep counterfactual abstraction. The current CoTs are mainly object-grounded next-subgoal statements: Figure~\ref{cot-self-correct} shows implicit condition-on-current-state re-planning after an execution error, but explicit branching or hypothetical simulation (e.g., ``if the grasp fails, then re-grasp'') remains future work.

    \subsection{Qualitative Case Study}
    \label{sec:appendix_casestudy}

    To qualitatively assess how explicit reasoning influences execution behavior, we compare rollout trajectories of DeepThinkVLA and the $\pi_0$-FAST baseline on representative LIBERO tasks (Figure~\ref{cot-self-correct}).
    The baseline policy initially approaches the target object but fails to establish a stable grasp.
    Following this error, it repeatedly executes similar motions without corrective adjustment, eventually leading to task failure.
    This behavior reflects a key limitation of direct perception-to-action policies: once an error occurs, the policy lacks an internal mechanism to reassess progress or revise its plan.

    In contrast, DeepThinkVLA generates a concise \texttt{<think>...</think>} trace prior to each action chunk.
    When an accidental drop occurs, the model explicitly re-articulates the subgoal (e.g., ``the butter needs to be moved into the basket''), enabling corrective re-grasping and successful task completion.
    This qualitative difference illustrates that CoT in DeepThinkVLA is not merely descriptive, but functions as an actionable planning scaffold that supports recovery and re-planning.
    As a result, DeepThinkVLA exhibits more stable and resilient behavior under execution errors and perturbations.

    \subsection{Detailed Ablation: SFT vs.\ RL on RoboTwin 2.0}
    \label{sec:appendix_robotwin_rl}

    To further substantiate the role of outcome-based reinforcement learning in complex, long-horizon settings, we provide a detailed comparison between SFT-only and RL-aligned models on the RoboTwin~2.0 benchmark.
    Table~\ref{tab:robotwin_rl} reports per-task success rates across short, medium, and long (including extra-long) horizons.

    The results show that RL yields consistent performance improvements across all horizon categories.
    Notably, the gains are most pronounced in medium (+9.5\%) and short (+7.5\%) horizon tasks, while still providing measurable benefits in long-horizon scenarios.
    These findings confirm that outcome-based RL plays a crucial role in refining reasoning--action alignment beyond what can be achieved through supervised imitation alone, particularly in high-fidelity, contact-rich environments.

    \begin{table}[t!]
    \centering
    \scriptsize
    \setlength{\tabcolsep}{8pt}
    \renewcommand{\arraystretch}{1.0}
    \caption{\textbf{RL vs.\ SFT on RoboTwin 2.0.}
    RL yields an overall \textbf{+6.8\%} gain (52.5\% $\to$ 59.3\%), confirming that outcome-based alignment is essential in contact-rich settings.}
    \label{tab:robotwin_rl}
    \resizebox{\linewidth}{!}{
    % \rowcolors*{1}{yellow!100}{yellow!100}
    \begin{tabular}{lccccc}
    \toprule
    \multicolumn{6}{c}{\textbf{Short Horizon Tasks (100--130 Steps)}} \\
    \midrule
    \textbf{Model} & \textbf{Lift Pot} & \textbf{Beat Hammer Block} & \textbf{Pick Dual Bottles} & \textbf{Place Phone Stand} & \textbf{Avg} \\
    \midrule
    DeepThinkVLA(SFT) & 61.0 & 64.0 & 46.0 & 19.0 & 47.5 \\
    DeepThinkVLA(RL) & 62.0 & 73.0 & 61.0 & 24.0 & 55.0 \\
    $\Delta$     & \textcolor{red}{+1.0} & \textcolor{red}{+9.0} &
                            \textcolor{red}{+15.0} & \textcolor{red}{+5.0} &
                            \textcolor{red}{+7.5} \\
    \midrule
    \multicolumn{6}{c}{\textbf{Medium Horizon Tasks (150--230 Steps)}} \\
    \midrule
    \textbf{Model} & \textbf{Move Can Pot} & \textbf{Place A2B Left}
                        & \textbf{Place Empty Cup} & \textbf{Handover Mic} & \textbf{Avg} \\
    \midrule
    DeepThinkVLA(SFT) & 33.0 & 30.0 & 78.0 & 82.0 & 55.8 \\
    DeepThinkVLA(RL) & 52.0 & 38.0 & 83.0 & 88.0 & 65.3 \\
    $\Delta$     & \textcolor{red}{+19.0} & \textcolor{red}{+8.0} &
                            \textcolor{red}{+5.0} & \textcolor{red}{+6.0} &
                            \textcolor{red}{+9.5} \\
    \midrule
    \multicolumn{6}{c}{\textbf{Long (280--320 Steps) \& Extra Long Horizon Tasks (450--650 Steps)}} \\
    \midrule
    \textbf{Model} & \textbf{Handover Block} & \textbf{Stack Bowls Two}
                        & \textbf{Blocks Rank Rgb} & \textbf{Put Bottles Dustbin} & \textbf{Avg} \\
    \midrule
    DeepThinkVLA(SFT) & 40.0 & 59.0 & 76.0 & 42.0 & 54.3 \\
    DeepThinkVLA(RL) & 43.0 & 62.0 & 77.0 & 49.0 & 57.8 \\
    $\Delta$     & \textcolor{red}{+3.0} & \textcolor{red}{+3.0} &
                            \textcolor{red}{+1.0} & \textcolor{red}{+7.0} &
                            \textcolor{red}{+3.5} \\
    \midrule
    \multicolumn{6}{l}{\textbf{Overall Gap ($\Delta$ = RL - SFT): \textcolor{red}{+6.8}}}\\
    \bottomrule
    \end{tabular}
    }
    \end{table}

    \subsection{Future Direction: Spatial Reasoning as CoT Augmentation}
    \label{appendix:spatial_reasoning}

    The CoT in DeepThinkVLA takes the form of natural-language next-subgoal statements. A natural question for future work is whether \emph{explicitly spatial} reasoning---e.g., predicting the end-effector trajectory that realizes the next subgoal---can serve as a complementary, more grounded intermediate representation along the CoT-to-action pathway.

\begin{figure}[t!]
        \centering
        \includegraphics[width=0.99\linewidth]{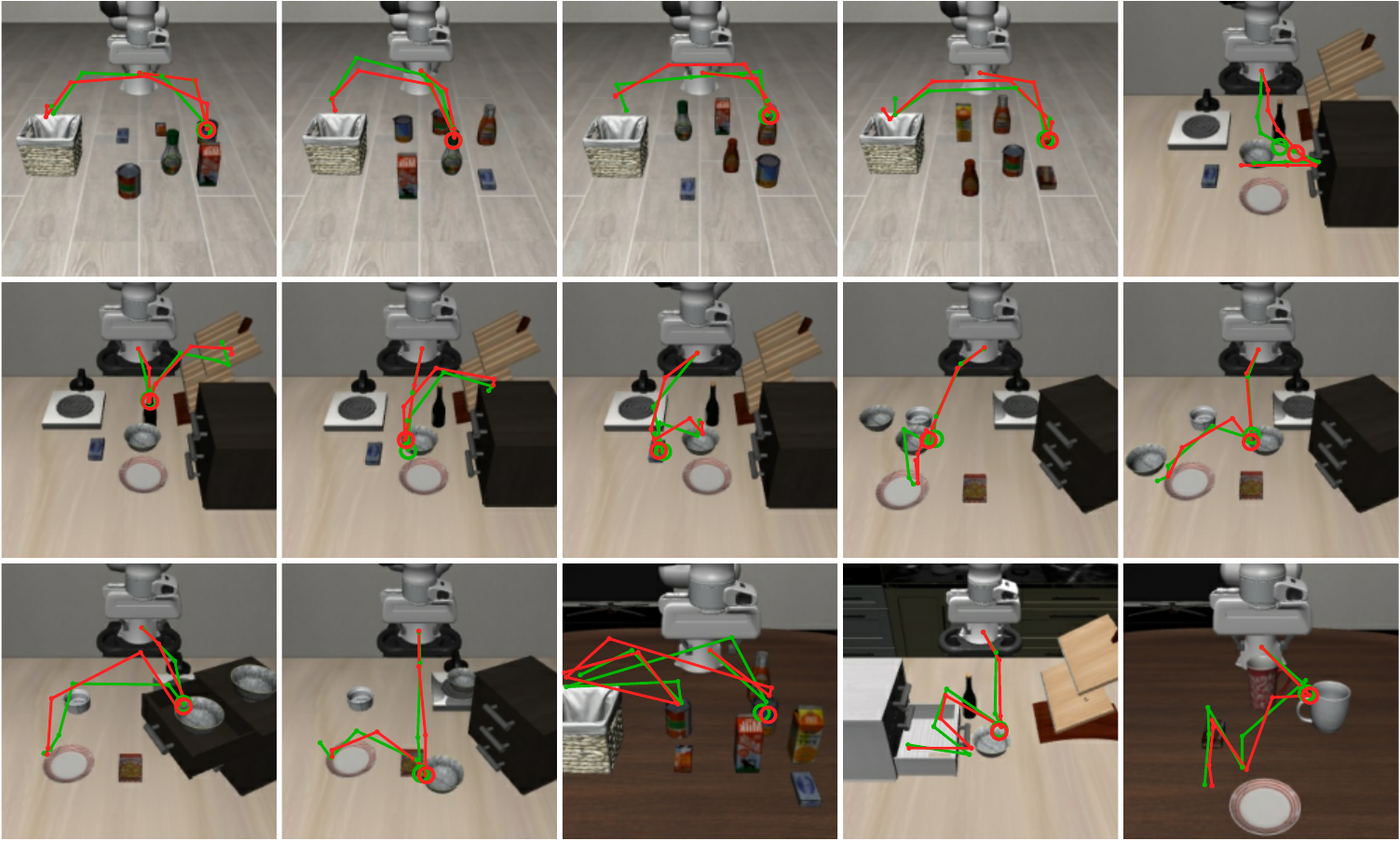}
        \caption{\textbf{Preliminary study: spatial-plan prediction with a LoRA-tuned RoboBrain~2.5 on LIBERO scenes.}
        Each panel overlays the ground truth (green) and the prediction of the LoRA-tuned RoboBrain~2.5 (red): lines trace the gripper waypoint path (end-effector positions sampled at action-chunk boundaries, projected into the image), and circles mark the grasp point (projected end-effector position at the first gripper closure). Closer red-to-green agreement indicates a more accurate spatial plan. The predictions are object-directed and broadly consistent with the ground truth, suggesting that spatial reasoning is a promising complement to language CoT.}
        \label{fig:spatial_grid}
    \end{figure}

    As a preliminary exploration of this direction, we conduct an experiment based on RoboBrain~2.5~\citep{tan2026robobrain}, an embodied vision-language foundation model with strong spatial reasoning and trajectory-generation capabilities. We fine-tune RoboBrain~2.5 with LoRA on spatial plans derived from our LIBERO demonstrations. For each episode, the ground-truth plan consists of (i) the \emph{gripper waypoint path}: the end-effector positions sampled at action-chunk boundaries (every 10 steps, yielding 8 waypoints), projected into the image and connected as a polyline; and (ii) the \emph{grasp point}: the projected end-effector position at the first gripper closure. Given the task instruction and the scene observation, the LoRA-tuned model predicts the same representation---a waypoint path and a grasp point in image space. Figure~\ref{fig:spatial_grid} presents qualitative results on 15 LIBERO scenes, overlaying the ground-truth plans (green) with the model predictions (red); the more closely the red path and circle track their green counterparts, the more accurate the predicted spatial plan.

    The predicted waypoint paths are generally object-directed and broadly consistent with the ground-truth paths across scenes and viewpoints, and the predicted grasp points fall on or near the target objects. This suggests a concrete future direction: augmenting the language CoT with a spatial reasoning channel---e.g., conditioning the parallel action decoder on both the generated subgoal text and a predicted spatial plan (waypoint path and grasp point)---so that the reasoning segment conveys not only \emph{what} to do next but also \emph{where} and \emph{how} to move. Integrating such spatially grounded CoT into the SFT-then-RL pipeline, and testing whether it further strengthens Causal Alignment under action-execution-sensitive OOD, is left to future work.

    We emphasize the scope of this study: it is an exploratory, qualitative experiment built on a separate model (RoboBrain~2.5 with LoRA adaptation) rather than on the DeepThinkVLA policy evaluated in the main paper, and none of the claims, conditions, or quantitative results in the main paper depend on it.

\end{document}